\documentclass[journal]{IEEEtran}
\usepackage{cite}

\ifCLASSINFOpdf
  \usepackage[pdftex]{graphicx}
  \graphicspath{{./pdf/}{./jpeg/}{.}}
  \DeclareGraphicsExtensions{.pdf,.png}
\else
\fi
\usepackage{bm}
\usepackage[cmex10]{amsmath}
\usepackage{amssymb}
\usepackage{cases}
\usepackage{gensymb}
\usepackage{soul,color,xcolor}
\newcommand{\tr}[1]{#1^{\mathsf{T}}}

\usepackage{array}

\ifCLASSOPTIONcompsoc
  \usepackage[caption=false,font=normalsize,labelfont=sf,textfont=sf]{subfig}
\else
  \usepackage[caption=false,font=footnotesize]{subfig}
\fi

\ifCLASSOPTIONdraftcls
  \usepackage[nomarkers]{endfloat}
\fi

\begin{document}
%
\title{Semi-Supervised Endmember Identification\\In Nonlinear Spectral Mixtures\\Via Semantic Representation}
%
%
%

\author{Yuki~Itoh,~\IEEEmembership{Student~Member,~IEEE,}
	Siwei~Feng,~\IEEEmembership{Student~Member,~IEEE,}
	Marco~F.~Duarte,~\IEEEmembership{Senior~Member,~IEEE,}
	and~Mario~Parente,~\IEEEmembership{Senior~Member,~IEEE}
	\thanks{\textcopyright\hspace{4pt}2017 IEEE. Personal use of this material is permitted. However, permission to use this material for any other purposes must be obtained from the IEEE by sending a request to pubs-permissions@ieee.org.}%
	\thanks{The authors are with the Department of Electrical and Computer Engineering, University of Massachusetts, Amherst, MA, 01003, USA. E-mail: \{yitoh, siwei\}@umass.edu and \{mduare, mparente\}@ecs.umass.edu.}%
	\thanks{This work was supported by the National Science Foundation under grant number IIS-1319585.}}%
\maketitle

\begin{abstract}
This paper proposes a new hyperspectral unmixing method for nonlinearly mixed hyperspectral data using a semantic representation in a semi-supervised fashion, assuming the availability of a spectral reference library. Existing semi-supervised unmixing algorithms select members from an endmember library that are present at each of the pixels; most such methods assume a linear mixing model. However, those methods will fail in the presence of nonlinear mixing among the observed spectra. To address this issue, we develop an endmember selection method using a recently proposed semantic spectral representation obtained via non-homogeneous hidden Markov chain (NHMC) model for a wavelet transform of the spectra. The semantic representation can encode spectrally discriminative features for any observed spectrum and, therefore, our proposed method can perform endmember selection without any assumption on the mixing model. Experimental results show that in the presence of sufficiently nonlinear mixing our proposed method outperforms dictionary-based sparse unmixing approaches based on linear models.
\end{abstract}


%
\IEEEpeerreviewmaketitle

\section{Introduction}
%
%
%
%

%
%
%
%
\IEEEPARstart{S}{pectral} unmixing aims at identifying the pure spectral signatures (endmembers) of each mixed pixel and estimating their fractional abundances collected by an imaging spectrometer. This process is crucial to inferring the compositions on the surface of Earth and planetary surfaces because the typical spatial resolution of hyperespectral images acquired by satellites is relatively large; therefore, each pixel is likely to be composed of multiple materials.

Researchers have addressed this issue by exploiting mixture models, both linear and nonlinear, to identify the endmembers composing each mixture~\cite{Bioucas-Dias2012JSTARS,Heylen2014}. The linear mixing model (LMM) assumes that the observed signature is approximated by a linear combination of the endmembers. Nonlinear mixing models (NLMM) have been proposed to deal with the fact that several aspects of the physical measurement of spectral signals introduce nonlinearities in the mixing process. In the NLMM, microscopic mixtures like intimate mixture or the multiple scattering effects at larger scale can be taken into consideration~\cite{Nascimento2009,Fan2009,Halimi2011,Altmann2012,Chen2013,Hapke2012,Close2012,Heylen2014a}. Several models to represent nonlinear mixing processes have been proposed. Some mixture models express the multiple scattering using an element-wise product of endmembers~\cite{Nascimento2009,Fan2009,Halimi2011,Altmann2012} or additive components~\cite{Chen2013}, while others use an intimate mixing using linear mixtures of albedos of endmembers~\cite{Hapke2012}. More recently, the hybrid of linear and intimate mixture models is also proposed to deal with more complex situations~\cite{Close2012,Heylen2014a}.

In terms of availability of an endmember dictionary, unmixing methods can be classified as supervised, unsupervised, and semi-supervised~\cite{Bioucas-Dias2012JSTARS}. The simplest scenario is the supervised one where the endmembers in the scene are assumed to be known a priori and only abundances are estimated. Unsupervised unmixing, where endmembers need to be estimated, has been also widely investigated in the literature~\cite{Bioucas-Dias2012JSTARS}. In contrast, semi-supervised unmixing is the most recently proposed approach that takes advantage of a spectral library that contains pure spectral signatures collected on the ground by a field spectrometer, or measured in the laboratory~\cite{Bioucas-Dias2010,Iordache2011TGRS,Iordache2011a,Iordache2012,Iordache2014,Iordache2014a,Themelis2012,Li2012,Shi2014,Tang2014,Tang2014a,Tang2015,Feng2015,Halimi2015,Zhang2016}. The spectral signatures are considered to be potential endmembers in the scene and hyperspectral unmixing is reduced to selecting the optimal combination of the pure signatures in the library that best model each mixed pixel in the scene. The spectral samples in the library usually correspond to well-characterized specific materials; therefore, we can directly perform material identification in the semi-supervised case, while unsupervised mixing requires the additional step of identifying the type of materials for each of the extracted endmembers.

Sparsity-based unmixing is perhaps the most popular semi-supervised method~\cite{Bioucas-Dias2010,Iordache2011TGRS}, receiving significant attention in recent years~\cite{Iordache2011a,Iordache2012,Iordache2014,Iordache2014a,Themelis2012,Li2012,Shi2014,Tang2014,Tang2014a,Tang2015,Feng2015,Halimi2015,Zhang2016}. However, most of the existing studies rely on the assumption of LMM, which is inappropriate if non-negligible nonlinear mixing is present. In fact, sparse unmixing (SU) based on LMM has been theoretically and experimentally shown to deteriorate its performance as the degree of nonlinearity in the mixing process increases~\cite{Yuki2015Whispers,Itoh2017TSP}. The previous study on SU with non-linear mixing models is limited to~\cite{Qu2014}, which only deal with a class of bilinear models.
On the other hand, choosing an appropriate NLMM can be a daunting task, since in one hyperspectral image scene it is likely that several different mixing phenomena are observed and therefore it would be appropriate to select models individually for each pixel.
Furthermore, even if one could pinpoint the exact nonlinear model to be applied, the complexity of inverting such a model can easily become untenable. For instance, any Hapke mixing model~\cite{Hapke2012} will cause a significant increase of the parameter space such as grain sizes, angles, backscattering functions, and phase functions. Additionally, a family of bilinear mixing models consider all combinations of two potential endmembers in the scene. 

In order to avoid the difficulties involved in choosing and inverting a model for nonlinear mixing, we propose a model-independent semi-supervised approach to endmember selection based on detecting endmember discriminative features that persist in mixed spectra regardless of the type of mixing present in the scene. Practitioners have attempted to perform identification of materials in the spectral library by relying on {\em semantic} features, which are associated with the chemical makeup of materials and observed as the specific position and shape of absorption bands in the spectral signals.~\cite{Clark1990a, Tetracorder,Swayze2014cuprite}. Here we define ``semantic'' features as ones that characterize a spectral signal to clearly differentiate them from ``discriminative'' ones. Those semantic features are routinely manually defined by experts are further used to determine discriminative spectral features, e.g., the Tetracorder algorithm~\cite{Tetracorder}.

We have recently proposed a semantic representation for hyperspectral signatures~\cite{Parente2013Whispers,Duarte2013Allerton}, which is obtained by modeling the wavelet coefficients of the signature with a non-homogeneous hidden Markov chain (NHMC) model. The NHMC model is shown to improve the performance of spectral classification when compared to competitor methods~\cite{Parente2013Whispers,Duarte2013Allerton}. The model is further explored in~\cite{Feng2014,Feng2015b,Feng2016}, where a more complex model to capture different magnitudes of fluctuations is proposed. This semantic representation can be effectively adopted to the unmixing problem, and particularly to the detection of endmembers.

This paper proposes an unmixing method based on the NHMC-based semantic representation. We have previously attempted NHMC-based unmixing by leveraging an early version of the statistical model~\cite{Itoh2014}. In contrast to this prior work, we now consider a more sophisticated NHMC model with multiple states; furthermore, we provide improvements on each of the stages of NHMC unmixing shown in Figure~\ref{fig:pipeline}. One of the biggest advantages of NHMC unmixing is that it makes no assumption on the spectral mixing model, because the semantic representation is independent of the type of spectral mixture. Hence, NHMC unmixing can be used for semi-supervised unmixing even when several nonlinear mixing phenomena are present.

In our framework, we cast the semi-supervised unmixing problem as an endmember selection problem and solve it by designing a series of detectors, each of which determines the presence of each endmember spectrum in a mixed pixel by using the semantic representation of the endmember spectrum. Therefore, instead of calling it NHMC unmixing, we call our proposed method {\it NHMC based endmember detector (NHMC-ED)} in the sequel. This approach relies on the observation that endmember semantic features persist in mixed spectra, albeit with attenuations, if the features for different endmembers don't overlap. Therefore, detecting such features in the mixed spectra could be a discriminative criterion for the presence of the endmembers. On the other hand, weak endmember features could be attenuated in the mixture. Furthermore, endmember features might be shared by more than one endmember family, including those not present in the mixture. Our approach relies on extracting a large number of features from an expanded library to address the attenuation problem and on a custom feature selection method to extract feature sets that are exclusive to each endmember and therefore truly discriminative.

We also note that our attention here is focused on endmember identification, and not on abundance estimation. While this could be seen as a limitation with respect to other approaches, we emphasize that correct selection of endmembers is perhaps the most important aspect of semi-supervised unmixing. Once we determine the endmembers that are present in the pixel, we can use the corresponding mixing model to estimate their abundances at a fraction of the computational complexity of existing approaches for joint unmixing and abundance estimation. 
As a matter of fact, accurate abundance estimation is a difficult problem in nonlinear unmixing due to the difficulty of collecting adequate ground-truth information on the abundance of endmembers contributing to a pixel in any non-trivial scenario. 

The contributions of our paper can be summarized as follows. We develop a new unmixing method using the semantic representation based on the recently proposed NHMC model; some novel aspects of our method include: ($i$) addressing the unmixing problem as a series of hypothesis testing problems, ($ii$) augmenting the spectral library to address the attenuation issue, and ($iii$) tailoring a new custom feature selection, based on conditional mutual information (CMI)~\cite{Fleuret2004}, to the unmixing problem. We also provide an extensive performance analysis of NHMC-ED and SU on both simulation and experimental data that provides interesting insights. The first and the second contributions are partly discussed in our previous work~\cite{Itoh2014} and their benefits are further investigated in this paper. In addition, we consider feature selection part in this paper in a more principled way.

The rest of the paper is organized as follows. Section~\ref{sec:semanticRepresentation} gives background on semantic representations based on the NHMC model. Section~\ref{sec:methodology} illustrates the proposed approach for unmixing, including a feature selection method for semantic features. Section~\ref{sec:experiment1} and~\ref{sec:experiment2} are devoted to experiments with synthetic and real data, respectively, and Section \ref{sec:conclusion} concludes the paper.

\section{Semantic representation of hyperspectral data}
\label{sec:semanticRepresentation}
In this section, we briefly describe the semantic representation of hyperspectral signals proposed in~\cite{Parente2013Whispers,Duarte2013Allerton,Feng2015b}. This representation is automatically generated by statistical analysis on the wavelet coefficients of the reflectance signals. We first describe the use of wavelet transforms for the spectral data, followed by a description of the statistical model placed on the wavelet coefficients, and end with a description of a semantic representation.

\subsection{Wavelet analysis of reflectance spectra}
The wavelet transform (WT), a popular tool for signal analysis, decomposes a signal into a multiscale time-frequency representation at different scales and offsets. The WT can be used for spectral signature analysis by associating WT components with absorption features in a spectrum that are routinely leveraged by experts. In fact, the WT itself has been used for classification of hyperspectral data~\cite{Zhang2005a,Masood2008,Prasad2012,Rivard2008}.

In particular, we use the undecimated wavelet transform (UWT) to detect localized features. The UWT encodes the magnitude of the convolution of a signal with a wavelet mother function at each wavelength without down-sampling for large scales; therefore, each convolved coefficient is considered to represent the power of the signal at a certain offset and scale and visually preserves the wavelength position of the features. In contrast, the decimated wavelet transform (DWT) creates a non-redundant feature obscuring the position of its entries due to the down-sampling of the signals.

In general, the UWT of a signal $\bm{y} = \tr{[y_1,y_2,\ldots,y_L]}$, where $L$ is the number of elements of the vector $\bm{y}$, is a convolution of a signal with a wavelet mother function at different scales,
\begin{equation}
w_{s,l} = \langle \bm{y}, \psi _{s,l} \rangle 
\end{equation}
where $w_{s,l} $ denotes a wavelet coefficient at a scale $s$ and an offset $l$, and $\psi(\cdot)_{s,l}$ denotes the mother function of the WT dilated to a scale $s$ and translated to an offset $l$, given by
\begin{equation}
\psi _{s,l}\left(\lambda \right) = \frac{1}{\sqrt{s}} \psi\left(\frac{\lambda - l}{s}\right).
\end{equation} 

\subsection{Statistical modeling on wavelet coefficients}
The WT encodes the signals in an energy-compact fashion, which makes the distribution of wavelet coefficients heavy tailed with a peak around zero. Those distributions can be modeled by a mixture of two Gaussians, both centered at the origin~\cite{Chipman1997}. One Gaussian in this mixture is assumed to have a small variance associated with the distribution of noise, and the other is assumed to have a large variance associated with the distribution of signal components.  Crouse et al.~\cite{Crouse1998} refined the probabilistic model by considering two observations: the persistence property, which addresses the propagation of large and small coefficients across scales, and the consistency property, which addresses the similarity of neighboring wavelet coefficients. These two properties motivate the construction of a hidden Markov tree (HMT) across the scales, which has been successful in modeling the signal's DWT coefficients that manifest themselves as a cone of influence in the wavelet coefficient matrix. The construction of a HMT enables us to mine signal components in small scales, which tend to be observed with small amplitudes. 

Inspired by this model, Duarte and Parente proposed a model on UWT coefficients that also present properties similar to those of the DWT~\cite{Parente2013Whispers,Duarte2013Allerton}. In this model, a non-homogeneous hidden Markov chain (NHMC) is constructed on the UWT coefficients across scales at each offset. As with~\cite{Chipman1997,Crouse1998}, the NHMC also models the distribution of each of the wavelet coefficients with a mixture of two zero-mean Gaussian distributions: with large or small variances, which are associated with two hidden states \{Large ($\mathsf{L}$), Small ($\mathsf{S}$)\} of the NHMC model. Recall that ($\mathsf{L}$) and ($\mathsf{S}$) indicate the presence or absence, respectively, of any fluctuation that is present in a signal at a specific location and scale. This hidden state of each wavelet coefficient has been used as a semantic representation that encodes the presence and location of signal fluctuations, and has been used to improve the accuracy of hyperspectral classification~\cite{Parente2013Whispers,Duarte2013Allerton}.

More recently, Feng et al.~\cite{Feng2015b} proposed a $k$-state mixture of Gaussian ($k$-MOG) NHMC model where the distribution of each wavelet coefficient is modeled as a $k$ Gaussian mixture model, with each Gaussian having mean zero and different variance. In this model, we also consider that semantic information is encoded in a binary fashion, \{Large ($\mathsf{L}$), Small ($\mathsf{S}$)\}, although this is not encoded directly into the hidden states. ($\mathsf{S}$) is assigned to a given wavelet coefficient if its hidden state is the one with smallest variance; otherwise, ($\mathsf{L}$) is assigned to the coefficient. We refer to this binary encoded semantic information \{($\mathsf{L}$), ($\mathsf{S}$)\} as a {\it feature label} to differentiate it from the MOG NHMC hidden state. The $k$-MOG NHMC model takes advantage of the granularity detected by the $k$-Gaussian mixture model  while reducing undesirable variation in models over all shifts and scales and is adopted in this paper.

We briefly review the $k$-MOG NHMC model proposed in~\cite{Feng2015b}. Each wavelet coefficient $w_{s,l}$ is assumed to be generated from one of the $k$ states denoted by $S_{s,l} \in \left\{ 0,1,\ldots,k-1 \right\}$, where the states have the prior probability $p_{s,l,i} = p(S_{s,l}=i)$ that meets the sum-to-one condition $\sum _i {p_{s,l,i}} = 1$. Let $S_{s,l}=0$ and $S_{s,l}>0$ correspond to ($\mathsf{S}$) and ($\mathsf{L}$) feature labels, respectively. In the sequel, we omit the subscript $l$ for the sake of simplicity (i.e., $w_{s,l} \rightarrow w_s$, $p_{s,l,i} \rightarrow p_{s,i}$, and $S_{s,l} \rightarrow S_s$). Each state is considered as a zero-mean Gaussian distribution expressed as 
\begin{equation}
w_{s}|S_{s}=i \; \sim \; \mathcal{N}\left(0,\sigma _{s,i}^2\right)
\end{equation}
where $\sigma _{s,i}^2$ is the variance for a state~$i$ at a scale~$s$ (specific to an offset~$l$). The marginal probability is computed by 
\begin{equation}p\left(w_{s}\right)=\sum _i {p_{s,i}\,p\left(w_{s}|S_{s}=i\right)}.
\end{equation}
The persistence property of the states across scales is modeled via a Markov chain on the hidden states of the UWT coefficients whose transition equation is given by
\begin{equation}
\bm{p}_{s+1} = \mathbf{A}_{s}\bm{p}_{s},
\end{equation}
where the vector of the state probabilities is defined by
\begin{equation}
\bm{p}_{s} = \tr{\left[ p_{s,0},\,p_{s,1},\ldots,p_{s,k-1}\right]}
\end{equation}
and the transition matrix of the state probabilities is defined as
\begin{equation}
\mathbf{A}_{s}=
\left[\begin{IEEEeqnarraybox*}[][c]{,c/c/c/c,}
p_{s,0\rightarrow 0} & p_{s,1\rightarrow 0} & \cdots & p_{s,k-1\rightarrow 0}\\
p_{s,0\rightarrow 1} & p_{s,1\rightarrow 1} & \cdots & p_{s,k-1\rightarrow 1}\\
\vdots & \vdots & \ddots & \vdots\\
p_{s,0\rightarrow k-1} & p_{s,1\rightarrow k-1} & \cdots & p_{s,k-1\rightarrow k-1}
\end{IEEEeqnarraybox*}\right]
,
\end{equation}
where $p_{s,i\rightarrow j}=p(S_{s+1}=j|S_{s}=i)$ expresses the transitional probability from a state $i$ to a state $j$ when we move from a scale $s$ to a scale $s+1$. Note that the diagonal elements of the transitional matrix $\mathbf{A}_{s}$ have larger values than others so that the persistence property across the scales holds. Note also that when almost all of the spectra have a fluctuation at a certain wavelength, the model learns a large variance for the state associated with the ($\mathsf{S}$) feature label, indicating that the features at that wavelength are nondiscriminative.

The $k$-MOG NHMC is independently trained on each different offset, namely on each of the $L$ wavelengths of the reflectance data, using a training set of spectra. The set of $k$-MOG NHMC parameters for the model at a given offset $l$ is defined as 
\begin{equation}
\Theta _l = 
\left\{\mathbf{A}_{s,l},\sigma_{s,l,1}^2,\ldots,\sigma _{s,l,k-1}^2 | s=1,\ldots,n_s \right\},
\end{equation}
where $n_s$ is the number of scales. The training of the model is performed via an expectation maximization algorithm that maximizes the expected log likelihood on the probabilistic distribution of the latent variables and states given a training dataset of spectra~\cite{Feng2015b}. After the model is trained, to obtain $k$-MOG NHMC feature labels, all hidden states associated with the ($\mathsf{L}$) feature label are merged into one state and a Viterbi algorithm~\cite{Durand2004,Viterbi1967} is used to estimate the most possible sequence of feature labels~\cite{Feng2015b}.

\section{NHMC-ED}
\label{sec:methodology}
In this section, we introduce the proposed NHMC-ED algorithm that uses the semantic representation described in Section~\ref{sec:semanticRepresentation}. Assuming the availability of a library of candidate endmember spectra, we solve the unmixing problem in a semi-supervised fashion by conducting endmember selection from the library. 
In particular, our method considers the use of binary semantic feature labels (obtained from an NHMC model) for unmixing. These features have been previously used for classification tasks where the learning process returns only one label out of the classes considered~\cite{Parente2013Whispers,Duarte2013Allerton,Feng2016}. In the case of unmixing, there may exist more than one material in the sample being considered; therefore, it is desirable to allow us to detect multiple endmembers simultaneously. One way to solve this problem is to define mixture classes, as in the Tetracorder~\cite{Tetracorder}. However, it would be computationally intractable to cover all the possible mixture combinations. Therefore, we will take a different approach to deal with this problem by defining one independent detection problem for each endmember class that is included in the library. In other words, detectors are designed in a class-wise fashion to enable us to identify the presence of multiple endmembers in the observation via a series of binary hypothesis tests. 

\subsection{Overview}
\label{sec:over}
Figure~\ref{fig:pipeline} illustrates the schematic of NHMC-ED, which is composed of two stages. In the {\em learning} stage, detectors for materials in the library are trained using the spectral library and the NHMC model; in the {\em testing} phase, each observed spectrum is examined to evaluate the presence of endmembers in the library using each of the learned decision rules on the spectrum's semantic features. The learning stage involves the computation of the parameters for the NHMC models from all the spectral samples in the dictionary, regardless of their material class. After the NHMC models are trained, we augment the dictionary with attenuated versions of the available spectral samples. This augmentation is key in our approach: although a pixel may be exclusively composed of one endmember, most pixel spectra are likely to be formed by a mixture with other materials and the concentration of an endmember may or may not be significant. In this case, the semantic features discriminative of each endmember might not be as pronounced in the mixed spectrum as the ones extracted from the endmember's pure spectral signature. Considering that only pure spectral samples of materials are contained in the library, learning discriminative features using only the pure spectral samples may overlook the necessary discriminability to detect the presence of such attenuated features. In other words, in order to detect the presence of an endmember in a mixture, the detector also needs to be tuned for the possibility of lower-contrast version of endmember discriminative features. This motivates our augmentation of the spectral library with attenuated versions of each spectral sample. In our implementation, we consider the multiplicative attenuation of library spectra.
\ifCLASSOPTIONdraftcls
	\begin{figure*}
		\subfloat[Learning stage]{\includegraphics[width=469pt]{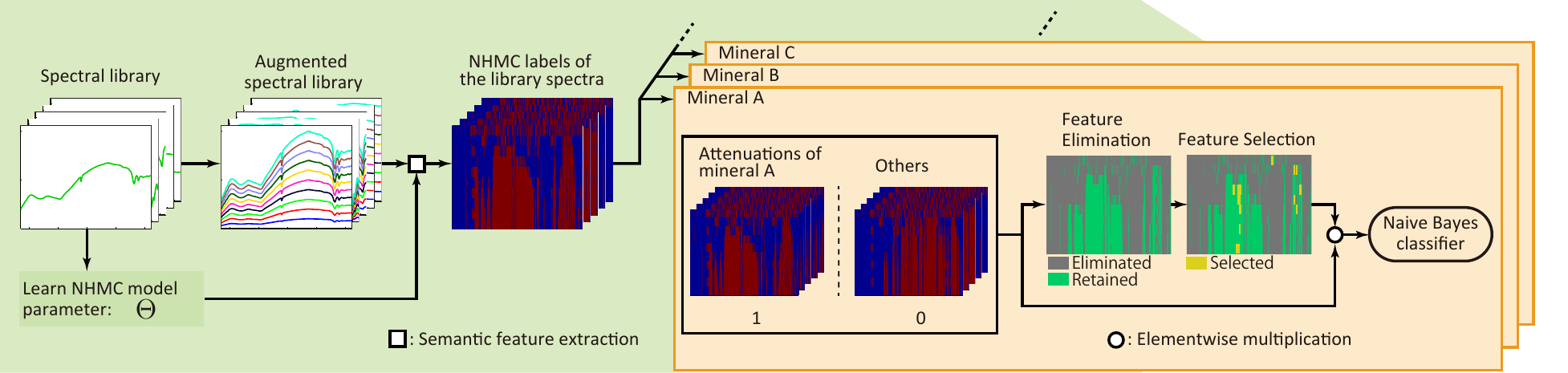}}\\
		\subfloat[Testing stage]{\includegraphics[width=469pt]{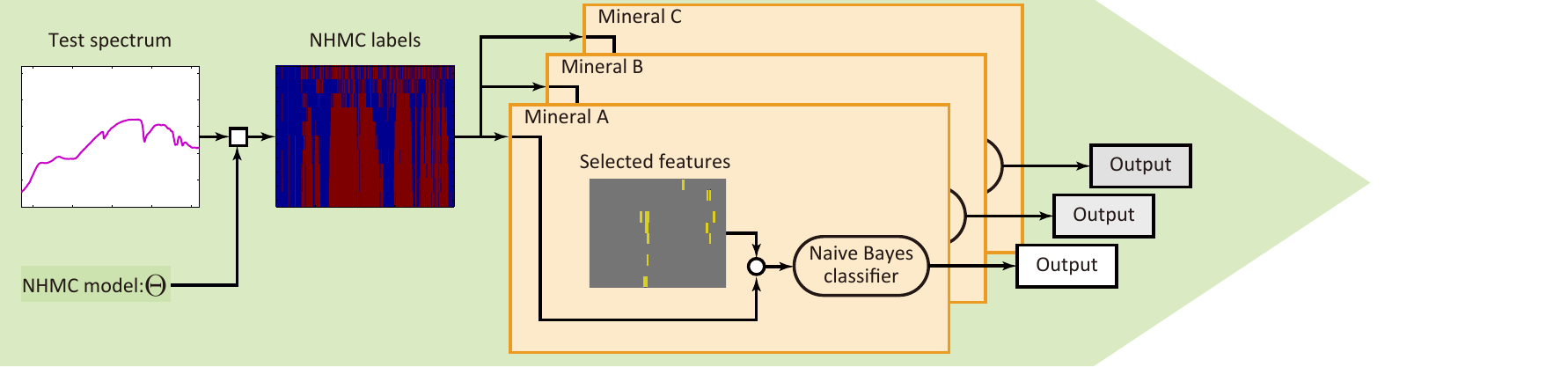}}\hfill
		\caption{A schematic of NHMC-ED, described in Section~\ref{sec:over}.}
		\label{fig:pipeline}
	\end{figure*}
\else
	\begin{figure*}
		\subfloat[Learning stage]{\includegraphics[width=181mm]{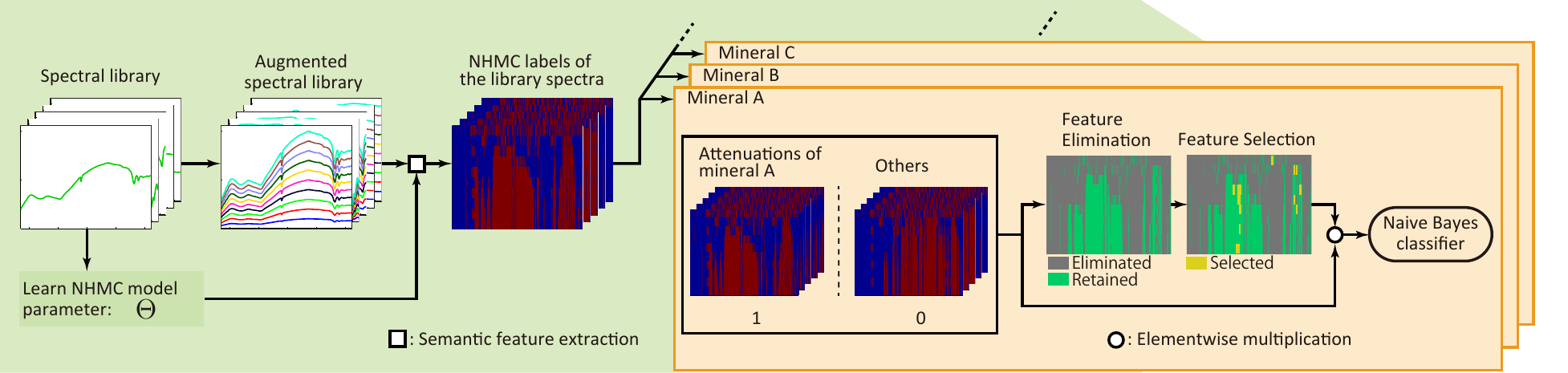}}\\
		\subfloat[Testing stage]{\includegraphics[]{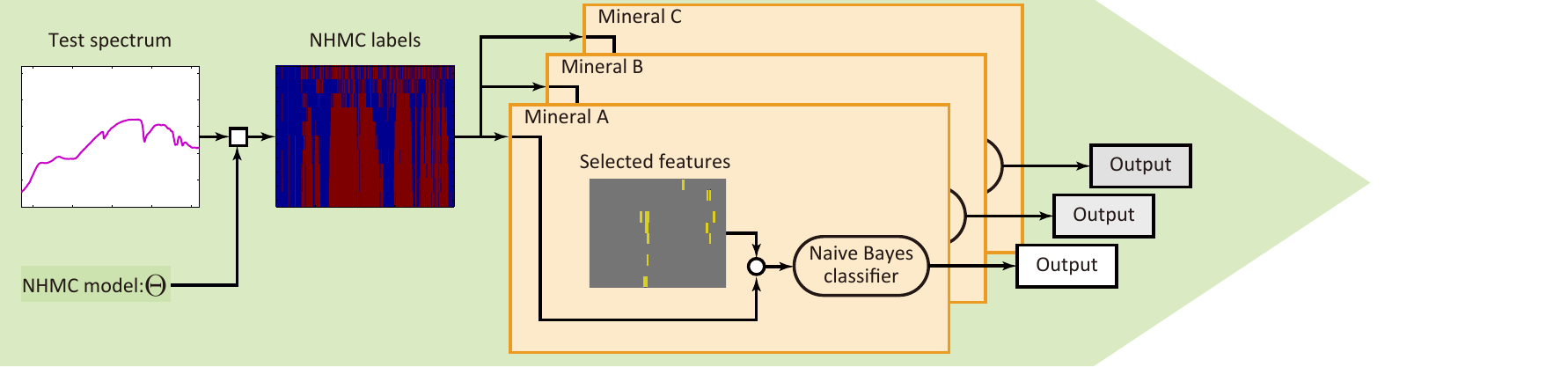}}\hfill
		\caption{A schematic of NHMC-ED, described in Section~\ref{sec:over}.}
		\label{fig:pipeline}
	\end{figure*}
\fi

After the library augmentation, the semantic representation of the spectral samples is obtained by computing the NHMC labels for the wavelet representation of those spectral signatures using the previously learned NHMC model parameters. The third column in Fig.~\ref{fig:pipeline}(a) shows some examples of the semantic representation encoded with binary NHMC state labels. The red and blue pixels indicate feature labels $(\mathsf{L})$ and $(\mathsf{S})$, respectively. 

Next, a detector is learned for each material class. The pink boxes (labeled Mineral A, Mineral B, Mineral C, etc.) in Fig.~\ref{fig:pipeline}(a) illustrate the detector learning for each material. In this stage, the spectral samples in the augmented library are first split into two classes: one is the set of samples of the material of interest (pure samples for the material class and their attenuated versions) and the other contains samples for all the other materials. Subsequently, discriminative feature sets that are exclusively discriminative of the material class of interest are extracted and used for its detection. This feature selection is composed of two steps: the pre-elimination of features and the feature selection on the retained ones. The rectangles labeled  ``$\mathsf{Feature~Elimination}$'' and ``$\mathsf{Feature~Selection}$'' portray the eliminated features (marked in green) and the selected discriminative features (marked in yellow), respectively. This procedure and the motivation of this two step approach is described in detail in Section \ref{sec:learnFeats}. Finally, a binary na\"{i}ve Bayes classifier is trained using only the selected features. 

In the testing stage, for each test spectrum, the NHMC state labels are computed using the model learned in the training stage, in order to obtain a semantic representation for the spectrum under test. Subsequently, the features selected for each detector are extracted individually and the corresponding previously trained na\"{i}ve Bayes classifiers are applied on the discriminative features to determine the presence of each of the materials in the library.

\subsection{Discriminating feature selection}
\label{sec:learnFeats}
We consider the use of a feature selection algorithm based on conditional mutual information (CMI)~\cite{Fleuret2004} and adapt it to our unmixing task. CMI-based feature selection attempts to select a set of features that are both maximally discriminant for the target variables and minimally redundant. In the algorithm, we iteratively and greedily add the unselected feature that maximizes the CMI, given the features that have been selected so far. 

Let $\bm{x} = \tr{[x_1,x_2,\ldots,x_N]}$ be the collection of binary feature labels of each sample spectrum over all scales and shifts, where $N$ is the number of features (in our specific case, $N$ is equal to the number of the wavelet coefficients), and $t$ be a scalar binary target variable, which represents the presence of the mineral of interest ($t=1$ and $t=0$ represent the presence and the absence of the material of interest, respectively), assigned to the feature vector $\bm{x}$. The mutual information between $\bm{x}$ and $t$ is given by
\begin{IEEEeqnarray*}[]{c}
	I(t;\bm{x}) = H(t) - H(t|\bm{x}),
\end{IEEEeqnarray*}
where $H(t)$ is the entropy of $t$ and $H(t|\bm{x})$ is the conditional entropy of $t$ given $\bm{x}$. CMI-based feature selection selects the least redundant subset of elements in $\bm{x}$ that can recover as much information of $t$ as all elements of $\bm{x}$ can. The feature selection method aggregates the most informative feature at each iteration in a greedy manner. Let the set of previously selected features at $i^{\mathrm{th}}$ iteration be $V_j=\{x_{v(1)},x_{v(2)},\ldots,x_{v(j)} \}$. At the $(j+1)^{\mathrm{th}}$ iteration the algorithm selects the feature that maximizes the CMI 
\begin{IEEEeqnarray*}[]{c}
	x_{v(i+1)} = \arg \max_{x_n} \; I(t;x_n|V_i),
\end{IEEEeqnarray*}
where $I(t;x_n|V_j)$ is the CMI between $t$ and $x_n$ given the feature set $V_i$, defined by
\begin{IEEEeqnarray*}[]{c}
	I(t;x_n|V_j) = H(t|V_j) - H(t|x_n,V_j).
\end{IEEEeqnarray*}

Under standard CMI feature selection, one searches for the features with highest correlation to the target label; such correlation can be {\em positive}  --- i.e. a feature is active ($\mathsf{L}$ is triggered) when the label is positive (1), and vice versa --- or {\em negative}  --- a feature is active when the label is negative (0), and vice versa. Unfortunately, negatively correlated features for single materials might become positive for mixtures that include that material if any other element of that mixture causes the feature to become active; this interference phenomenon can potentially affect the labeling for any  class that is negatively correlated with a feature. To address this issue, our feature selection will only focus on features that are positively correlated to the label.

To detect only positively correlated features, we define the error matrix
\begin{equation}
\mathbf{E} = \left[\begin{IEEEeqnarraybox*}[][c]{,c/c,}
p(0,0)&p(0,1)\\
p(1,0)&p(1,1)%
\end{IEEEeqnarraybox*}\right],
\end{equation}
where $p(x,t)$ denotes the joint distribution of a feature $x$ and its target label $t$, and both $x$ and $t$ are binary. The determinant of this error matrix is expressed as
\begin{equation}
\det{\mathbf{E}} = p(0,0)p(1,1) - p(0,1)p(1,0).
\end{equation}
Thus, if the determinant of this matrix is greater than zero, the diagonal elements are dominant, which means that the feature and its target are positively correlated. We eliminate features for which $\det{\mathbf{E}} \le 0$. After this feature elimination, we apply the CMI-based feature selection algorithm as usual.

\subsection{Detection of endmembers}
Testing the feature selection algorithm with various kinds of classifiers showed that the best overall classification performance was obtained by matching the CMI feature selection with a na\"{i}ve Bayes (NB) classifier~\cite{Fleuret2004}; we conjecture that this was achieved due to the fact that the  CMI-based approach tends to select features with as little dependency as possible, which is in line with the assumption of feature independence made in the design of the classifier.
For this reason, the detector of each endmember class is designed using a NB classifier. We denote a vector $\bm{x}_{V_K} = \tr{[x_{v(1)},x_{v(2)},\ldots,x_{v(K)}]}$  containing the $K$ selected binary features ($x_{v(j)} \in \{0,1\}$ for all $j$), and a binary target label variable $t\in \{0,1\}$. The NB classifier~\cite{PRML} assumes mutual conditional independence among the features, i.e.,
\begin{IEEEeqnarray}[]{r,c,l}
	p(\bm{x}_{V_K}|t) = \prod_{j=1}^{K}{p(x_{v(j)}|t)},
\end{IEEEeqnarray}
and evaluates the log-likelihood
\begin{IEEEeqnarray*}[]{r,c,l}
	&& \log{p(t=i|\bm{x}_{V_K})} \\
	&=& \log\left(p(\bm{x}_{V_K}|t=i)p(t=i)\right)+c\\
	&=& \sum_{j=1}^{K}\log\left(p(x_{v(j)}|t=i)\right)+\log\left(p(t=i)\right)+c, \IEEEyesnumber\label{eq:cond}
\end{IEEEeqnarray*}
where $c$ is a constant that does not depend on $i$. We can subsequently write $p(x_{v(j)}|t=i) = p_{v(j)i}^{x_{v(j)}}{(1-p_{v(j)i})}^{1-x_{v(j)}}$, where $p_{v(j)i} = p(x_{v(j)}=1|t=i)$. By substituting this into (\ref{eq:cond}), we obtain
\begin{IEEEeqnarray*}[]{r,c,l}
	\log{p(t=i|\bm{x}_{V_K})} &=& \sum_{j=1}^{K}x_{v(k)}\log{\frac{p_{v(j)i}}{1-p_{v(j)i}}} +\log (p(t=i))\\
	&&+\sum_{j=1}^{K}\log{(1-p_{v(j)i})}+c.
\end{IEEEeqnarray*}
In the training of the classifier, $p_{v(j)i}$ and $p(t=i)$ are learned by maximum likelihood estimation. In the testing stage, the estimated label $\widehat{t}$ is given by
\begin{IEEEeqnarray*}[]{c}
	\widehat{t} = \arg \max_{i\in\{0,1\}} \; \log{p(t=i|\bm{x}_{V_K})}.
\end{IEEEeqnarray*}

\subsection{Computational complexity of NHMC-ED}
As with most statistical approaches, the computational time of our method is dominated by the complexity of learning the statistical model, namely obtaining NHMC model parameters, which is the first step in the learning stage of our method shown in Fig.~\ref{fig:pipeline}(a). However, its computational complexity is difficult to assess because it is based on the EM algorithm. Here we describe the complexities, except for that step.

Recall that $L$ is the number of spectral channels, $k$ is the number of NHMC states, $n_s$ is the number of scales for the wavelet transform, and $K$ is the number of selected features. In addition to that, let $N_l$ be the number of elements in the library, $N_a$ be the number of observed spectral signals, $N_c$ be the number of classes in the library, $s_a$ be the ratio of the size of the augmented library to that of the library.

The learning stage is composed of five different steps. Table~\ref{table:compcmplx_learning} shows the computational complexity at each step. The total time complexity in the learning stage (except for the first step where the NHMC model parameters are learned) is equal to $\mathcal{O}((k^2+KN_c)n_sLN_ls_a)$.  Table~\ref{table:compcmplx_apply} shows the computational complexity at each spectrum identification step. The total time complexity in the learning stage is equal to $\mathcal{O}((k^2n_sL+KN_c)N_a)$. Therefore, the total time complexity is $\mathcal{O}((k^2+KN_c)n_sLN_ls_a)+(k^2n_sL+KN_c)N_a) = \mathcal{O}(k^2n_sL(N_ls_a+N_a)+KN_c(n_sLN_ls_a + N_a))$.

\begin{table}[]
	\renewcommand{\arraystretch}{1.4}
	\setlength{\extrarowheight}{1.5pt}
	\caption{Computational complexity in the learning stage}
	\label{table:compcmplx_learning}
	\centering	
	\begin{tabular}[c]{|l|l|}
		\cline{1-2}
		Augmentation of the library & $\mathcal{O}(LN_ls_a)$ \\
		\cline{1-2}
		Wavelet transform & $\mathcal{O}(n_sLN_ls_a)$ \\
		\cline{1-2}
		Label feature learning (Viterbi algorithm) & $\mathcal{O}(k^2n_sLN_ls_a)$ \\
		\cline{1-2}
		Negative feature elimination & $\mathcal{O}(N_cn_sLN_ls_a)$ \\
		\cline{1-2}
		CMI feature selection & $\mathcal{O}(KN_cn_sLN_ls_a)$ \\
		\cline{1-2}
		Learning Na\"{i}ve Bayes classifier & $\mathcal{O}(KN_cLN_ls_a)$ \\
		\cline{1-2}
	\end{tabular}
\end{table}

\begin{table}[]
	\renewcommand{\arraystretch}{1.4}
	\setlength{\extrarowheight}{1.5pt}
	\caption{Computational complexity in the testing stage}
	\label{table:compcmplx_apply}
	\centering	
	\begin{tabular}[c]{|l|l|}
		\cline{1-2}
		Wavelet transform & $\mathcal{O}(n_sLN_a)$ \\
		\cline{1-2}
		Label feature learning (Viterbi algorithm) & $\mathcal{O}(k^2n_sLN_a)$ \\
		\cline{1-2}
		Applying Na\"{i}ve Bayes classifier & $\mathcal{O}(KN_cN_a)$ \\
		\cline{1-2}
	\end{tabular}
\end{table}

\section{Experimental results with simulation data}
\label{sec:experiment1}
In this section, we test NHMC-ED with various kinds of mixture models, and compare its performance to that of the state-of-the-art SU approach. Our simulation uses a spectral database taken by the NASA RELAB facility at Brown University~\cite{relab}. We choose 360 samples with 14 classes such that the samples have reflectance data in the visible near-infrared region (300 -- 2600 nm) with 5 nm spectral resolution. The selected spectra were measured from particulate (powdered) samples obtained at several particle sizes. We randomly divide the available sample spectra into a training and testing data set, so as to represent the typically occurring scenario in which a different ``exponent" of a spectral class is present in a scene with respect to the sample acquired in the laboratory and included in the reference library, due to different acquisition conditions, possible presence of trace impurities and other environmental effects. This manipulation mimics the situation where spectral variability exists, providing a potential source of nonlinearity additional to the mixing process. The training and testing samples are divided so that they are as different to each other as possible. To achieve this differentiation, we perform $K$-means clustering with $K=2$ for each class; samples in one cluster are used for training, while samples in the other cluster are used for testing. After the data is partitioned, we generate additional training samples by mixing the training data within each mineral class using Hapke mixing~\cite{Hapke2012} in order to equalize the number of samples for each mineral class to 43 (i.e., to match the largest of the classes among the training set). 

The performance of NHMC-ED is compared with that of SU, which assumes a linear mixing model. In particular, we use spectral unmixing by splitting and augmented Lagrangian (SUnSAL)~\cite{Bioucas-Dias2010} and collaborative SUnSAL (CLSUnSAL)~\cite{Iordache2014}. The sum-to-one constraint of SUnSAL is discarded because this reportedly improves the performance~\cite{Iordache2011TGRS}.  Although they estimate the abundance of endmembers in the library, we only examine their detection performance. For this purpose, we further reject endmembers with sufficiently small abundances by hard thresholding. The coefficient on the regularization term and the threshold value are considered as the two parameters for each of SUnSAL and CLSUnSAL. 

\subsection{Performance metrics}
\label{sec:perfmetrics}
Since we are considering the detection of materials that are listed in the dictionary, it is natural to use standard metrics for target detection to assess the performance. Thus, we use the recall ($\mathrm{R}$) and false alarm rate ($\mathrm{FAR}$) metrics, defined by
\begin{IEEEeqnarray*}[]{c}
	\mathrm{R} = \frac{\mathrm{TP}}{\mathrm{TP}+\mathrm{FN}}, \,\,\, \mathrm{FAR} = \frac{\mathrm{FP}}{\mathrm{FP}+\mathrm{TN}},
\end{IEEEeqnarray*}
where $\mathrm{TP}$, $\mathrm{FN}$, $\mathrm{FP}$, and $\mathrm{TN}$ denote the number of true positives, false negatives, false positives, and true negatives, respectively. 
We use the receiver operating characteristic (ROC) curve to compare the performance of NHMC-ED and alternative unmixing methods. The ROC curve characterizes the performance of a binary detector and is drawn by plotting FAR and Recall in the $x$ and $y$ axis for different parameters of the detector. If the parameter space has more than one dimension, the ROC curve may be interpreted as a mesh. For brevity we will refer to the ROC curve/mesh as a curve. The ROC curve of NHMC is generated by varying the values of two parameters: the number of states in the NHMC model and the number of selected features. Similarly, the ROC curves of SUnSAL and CLSUnSAL are generated by varying the values of the trade-off parameter and the thresholding level. We define the performance of endmember detection by finding the closest distance $d_{\text{ROC}}$ between points in the ROC curve/mesh and the upper left corner of the ROC plot; more specifically,
\begin{IEEEeqnarray*}[]{c}
	d_{\text{ROC}} := \min \;\sqrt{\mathrm{(1-R)}^2 + \mathrm{FAR}^2}.
\end{IEEEeqnarray*}

Throughout this section, the search range of the two parameters for the NHMC-ED are as follows: the number of states for the NHMC model is set to $[2,4,6,8]$; the number of features retrieved by feature selection is set to $[1,2\ldots,50]$; and the range of the trade-off parameter is set to $[0,10^{-4},10^{-2},10^{-1}]$ for SUnSAL and to $[10^{-4},5\times10^{-4},10^{-2},10^{-1}]$ for CLSUnSAL. In the subsequent thresholding, 70 threshold values between $0$ and $1$ are tested. The number of scales used for the wavelet transform the proposed method is set to $n_s=10$.

\subsection{Construction of synthetic data}
We investigate the performance of NHMC-ED and SU with several mixing models that exhibit varying degrees of nonlinearity. In this experiment, we consider the LMM as well as several bilinear models --- Fan's model (FM), Nascimento's model (NM), generalized bilinear model (GBM), and polynomial post nonlinear model (PPNM) (cf.\ \cite{Dobigeon2014}, and references therein) --- and Hapke's model (HM)~\cite{Hapke2012}. We also includes an extreme case of NM that has only the second-order terms, which we call second-order-nonlinear model (SM). 

We introduce the {\em nonlinearity score} ($\mathrm{NS}$)~\cite{Itoh2016IGARSS} to measure the degree of nonlinearity exhibited by the mixed pixels, which corresponds to the angle between the nonlinearly mixed observation and the closest point in the convex cone defined by the endmembers involved; in other words, $\mathrm{NS}$ measures the size of the angle that the observed spectrum deviates from the set of all possible linear (conic) combinations of the endmembers, which can be computed as
\begin{IEEEeqnarray*}{c}
	\mathrm{NS}(\bm{y}_p) = \arccos \biggl( \frac{\tr{\bm{y}_p}\mathbf{W}\bm{a}^{*}(\bm{y}_p)}{\| \bm{y}_p \| \| \mathbf{W}\bm{a}^{*}(\bm{y}_p) \|} \biggr)\IEEEyesnumber\label{eq:NS}.
\end{IEEEeqnarray*}
Here $\mathbf{W}$ is a matrix whose columns are the endmember vectors from the training library for the classes involved in the mixture $\bm{y}_p$, and $\bm{a}^{*}(\bm{y}_p)$ is the best linear approximation to $\bm{y}_p$ over the endmembers in $\mathbf{W}$, given by
\begin{IEEEeqnarray*}[]{r'c}
	\bm{a}^{*}(\bm{y}_p) = & \arg  \min_{\bm{a}\succeq \bm{0}} \quad \|\bm{y}_p-\mathbf{W}\bm{a}\| _2
\end{IEEEeqnarray*}
Here $\bm{a}$ denotes an abundance vector.
Since $\bm{y}_p$ is generated from the testing library, the nonlinearity score also accounts for the deviation from linearity due to the mismatch between the training and testing libraries. Under the NS score defined above, even the LMM produces some points with deviations from linearity because different endmember sets are used in training and testing. On top of this mismatch distortion, the nonlinearity increases as the contribution of the weight to the second-order terms in bilinear models (NM, FM, GBM, and PPNM) increases. HM also produces some nonlinearity because it considers the linear mixing of SSAs that are created by the nonlinear conversion of the reflectance. 

We first construct synthetic mixture spectra using the nonlinear models listed above
with spectra from the testing library; recall that this library contains the same classes (but different samples for each class) as the training library. We first randomly select 50 different combinations from the $M$ endmember classes for each given value of $M$. For each combination, we select one endmember element for each class from the testing dataset (uniformly at random) to create a test mixture. These endmembers are mixed using the aforementioned mixing models under 500 different mixing weights, resulting in 25,000 spectra for each mixture type. The abundances of LMM, HM, FM, GBM, PPNM, and SM are drawn from the uniform distribution on the simplex, i.e., the Dirichlet distribution with parameter values being all equal to one. For NM, the abundances and coefficients for the second order terms are drawn from the uniform distribution on the simplex. Each of the coefficients that control the power of the second order term in GBM and PPNM is independently drawn from the uniform distribution over $(0,1)$ and $(-3,3)$, respectively as in~\cite{Halimi2011,Altmann2012}.
For HM, single scattering albedos (SSAs) for the endmembers are obtained by inverting the model using the recorded incident and emission angles. Assuming that the porosity parameter $K=1$, the phase function $p(g)=1$, and the back scattering function $B(g)=0$, the SSAs of the endmembers are computed by using the bisection method. The same assumption is made for $K$, $p(g)$, and $B(g)$ to construct mixtures using the SSAs of the endmembers.

Finally, we add zero-mean Gaussian noise to synthetically mixed spectra with a signal-to-noise ratio (SNR) of 50dB. The SNR [dB] is defined as $\log_{10}(\mathbb{E}[\tr{\bm{y}}\bm{y}]/\mathbb{E}[\tr{\bm{n}}\bm{n}])$ where $\mathbb{E}[\cdot]$ represents the expectation operator and $\bm{n}$ is a noise vector of length $L$. For given SNR, the variance of the noise is computed via $\mathbb{E}[\tr{\bm{y}}\bm{y}]/(10^{\mathrm{SNR}/10}L)$.

\subsection{Experimental results}
Before proceeding to our analysis, we set the default number of endmembers in the mixtures to $3$. The attenuations of library spectra range on a grid from 0.1 and 1 with a step size of 0.1.

We first investigate the impact of two new components of our proposed NHMC-ED: the library augmentation (LA) and negatively correlated feature elimination (NCFE) prior to CMI-based feature selection. Table~\ref{table:nhmc} shows $d_\mathrm{ROC}$ on the seven different mixing models for four methods: NB classifier directly on the feature label without LA or feature selection (labeled ``NB''), NB with NCFE and CMI-based feature selection (labeled ``NCFE+NB''), NB with LA and CMI-based feature selection (labeled ``LA+NB''), and NB with LA, NCFE, and CMI-based feature selection (labeled ``NCFE+LA+NB''). Overall, the proposed NHMC-ED, which uses both NCFE and LA, tends to perform best and show robustness to different mixing models. In addition, we can see the benefit of LA and NCFE by comparing the performance of ``NCFE+NB'' and ``LA+NB'' with the rightmost column, respectively. The numbers in the second column from the right show better performance than those in the third column from the right, indicating that LA is more critical to the performance than NCFE; however, both of them seem to contribute the improvement in performance. Compared to NB, the other methods consistently perform much better, indicating that semantic information retrieved by NHMC modeling should be combined with an appropriate feature selection method to achieve accurate endmember detection.

\begin{table}[b]
	\renewcommand{\arraystretch}{1.4}
	\setlength{\extrarowheight}{1.5pt}
	\caption{Performance of NHMC-ED with or without new components}
	\label{table:nhmc}
	\centering	
	\begin{tabular}[c]{|c|c|c|c|c|}
		\cline{2-5}
		\multicolumn{1}{c|}{} & \multicolumn{4}{c|}{$d_\mathrm{ROC}$} \tabularnewline
		\cline{1-5}
		Model & NB & NCFE+NB & LA+NB & NCFE+LA+NB \tabularnewline
		\cline{1-5}
		LMM  & 0.846 & 0.482 & 0.389 & {\bf 0.319} \tabularnewline
		\cline{1-5}
		GBM  & 0.834 & 0.521 & 0.442 & {\bf 0.360} \tabularnewline
		\cline{1-5}
		FM   & 0.853 & 0.437 & 0.344 & {\bf 0.308} \tabularnewline
		\cline{1-5}
		HM   & 0.865 & 0.384 & {\bf 0.351} & 0.360 \tabularnewline
		\cline{1-5}
		NM   & 0.843 & 0.500 & 0.422 & {\bf 0.344} \tabularnewline
		\cline{1-5}
		PPNM & 0.907 & 0.549 & 0.517 & {\bf 0.484} \tabularnewline
		\cline{1-5}
		SM   & 0.850 & 0.487 & 0.407 & {\bf 0.333} \tabularnewline
		\cline{1-5}
	\end{tabular}
\end{table}
\ifCLASSOPTIONdraftcls
\begin{figure}[t]
	\centering
	\includegraphics[]{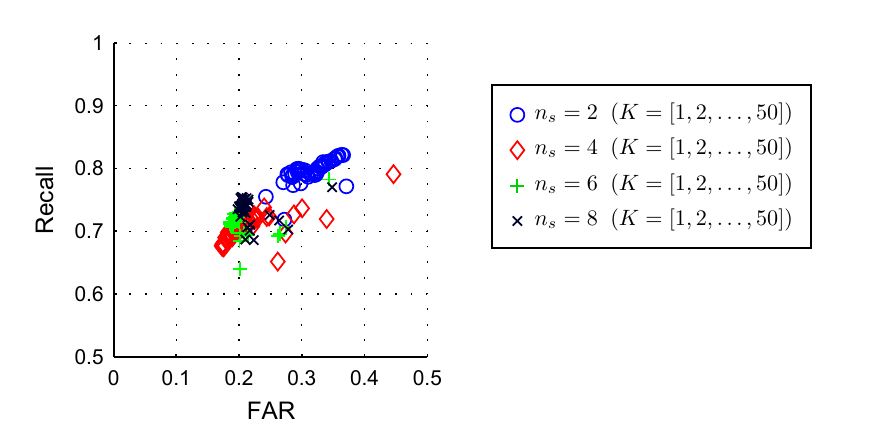}
	\caption{The recall/FAR performance plot of NHMC-ED NHMC-ED parameters, the number of scales $n_s$ and the number of features $K$. Each color and marker denotes a distinct value for $n_s$ while $K$ varies.}
	\label{fig:nhmcPrfParam}
\end{figure}
\else
\begin{figure}[t]
	\centering
	\includegraphics[]{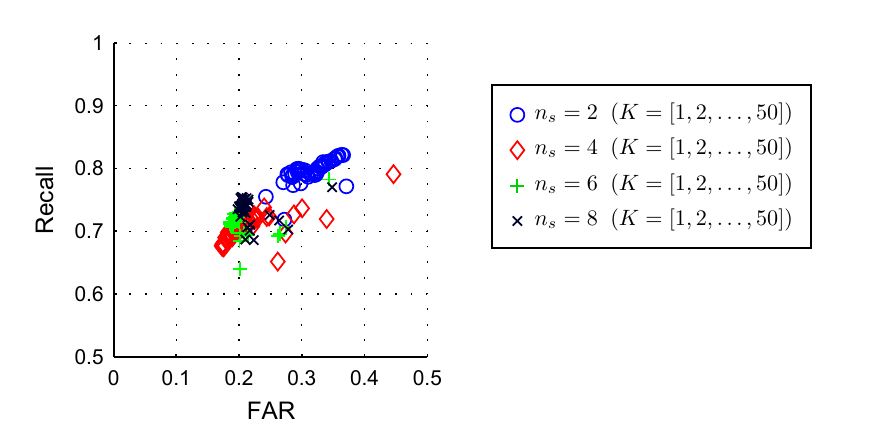}
	\caption{The recall/FAR performance plot of NHMC-ED for different parameters, the number of scales $n_s$ and the number of features $K$. Each color and marker denotes a distinct value for $n_s$ while $K$ varies.}
	\label{fig:nhmcPrfParam}
\end{figure}
\fi
Next, we investigate the sensitivity of NHMC-ED to different values of its parameters. Figure~\ref{fig:nhmcPrfParam} shows its performance mesh on HM for different number of states of the NHMC model and for different number of features selected by CMI-based feature selection. Each curve corresponds to a fixed number of states and connects the performance point for a varying number of features. From this figure, it can be seen that the variation in performance is small. Nonetheless, it can be challenging to predict the optimal parameter combination for NHMC.

Third, we compare the performance of our methods with the two SU approaches. From now on, NHMC-ED refers to our proposed approach using both NCFE and AL. Table~\ref{table:prfComp} shows the performance of NHMC-ED, SUnSAL, and CLSUnSAL on various kinds of mixture models in order of increasing average nonlinearity score $\mathrm{NS}(\bm{y})$. According to the table, NHMC-ED comes to perform best as the degree of nonlinearity increases. This result demonstrates the NHMC-ED performs better in the presence of a sufficiently strong degree of nonlinearity, indicating its robustness to non-linear mixing. CLSUnSAL tends to perform worse than SUnSAL. We conjecture that this is because the uniform mixing assumption made by CLSUnSAL  is not valid in these simulated data. Clustering of mixtures prior to unmixing~\cite{arun2016} may be necessary to take advantage of the model assumed by CLSUnSAL.

\begin{table}[b]
	\renewcommand{\arraystretch}{1.4}
	\setlength{\extrarowheight}{1.5pt}
	\caption{Performance of NHMC-ED and SU on different mixing models}
	\label{table:prfComp}
	\centering	
	\begin{tabular}[c]{|c|c||c|c|c|}
		\cline{3-5}
		\multicolumn{2}{c|}{} & \multicolumn{3}{c|}{$d_\mathrm{ROC}$} \tabularnewline
		\cline{1-5}
		Model & $\mathrm{NS}$ [deg] & SUnSAL & CLSUnSAL & NHMC-ED \tabularnewline
		\cline{1-5}
		LMM   & 1.357 & {\bf 0.238} & 0.268 & 0.360 \tabularnewline
		\cline{1-5}
		GBM   & 1.502 & {\bf 0.255} & 0.325 & 0.344 \tabularnewline
		\cline{1-5}
		FM   & 1.664 & {\bf 0.270} & 0.378 & 0.333 \tabularnewline
		\cline{1-5}
		HM   & 2.681 & 0.348 & 0.344 & {\bf 0.319} \tabularnewline
		\cline{1-5}
		NM   & 2.841 & {\bf 0.304} & {\bf 0.304} & 0.308 \tabularnewline
		\cline{1-5}
		PPNM   & 5.913 & 0.536 & 0.650 & {\bf 0.484} \tabularnewline
		\cline{1-5}
		SM   & 6.361 & 0.427 & 0.423 & {\bf 0.360} \tabularnewline
		\cline{1-5}
	\end{tabular}
\end{table}

\ifCLASSOPTIONdraftcls
\begin{figure}[t]
	\centering
	\includegraphics[]{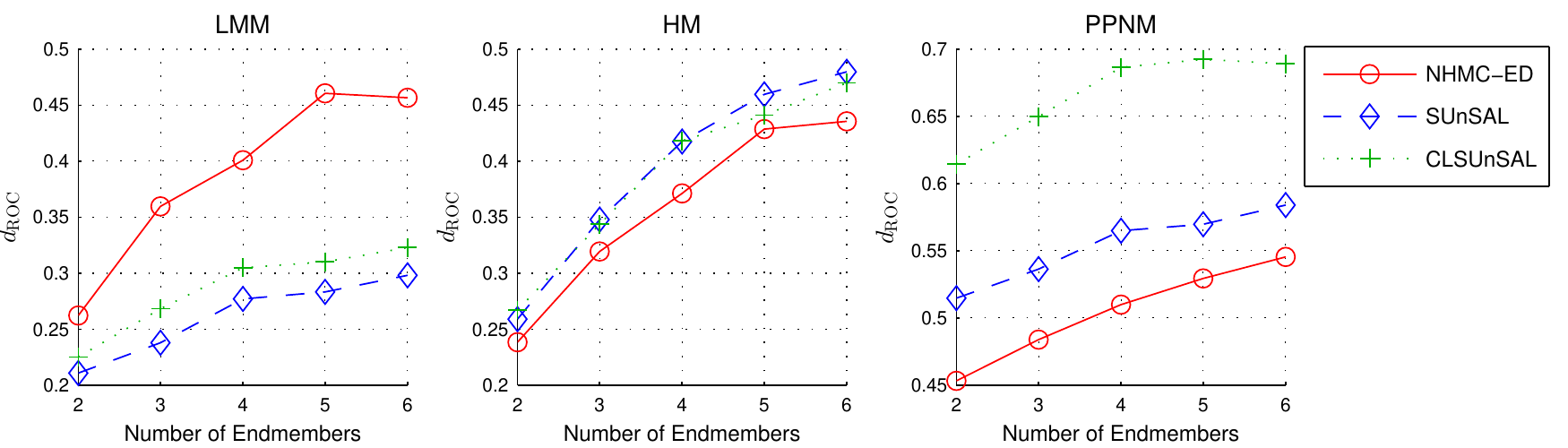}
	\caption{Comparison of the performance of NHMC-ED and SU approaches on synthetic mixtures as a function of different number of endmembers.}
	\label{fig:compPrfnEM}
\end{figure}
\else
\begin{figure*}[t]
	\centering
	\includegraphics[width=510pt]{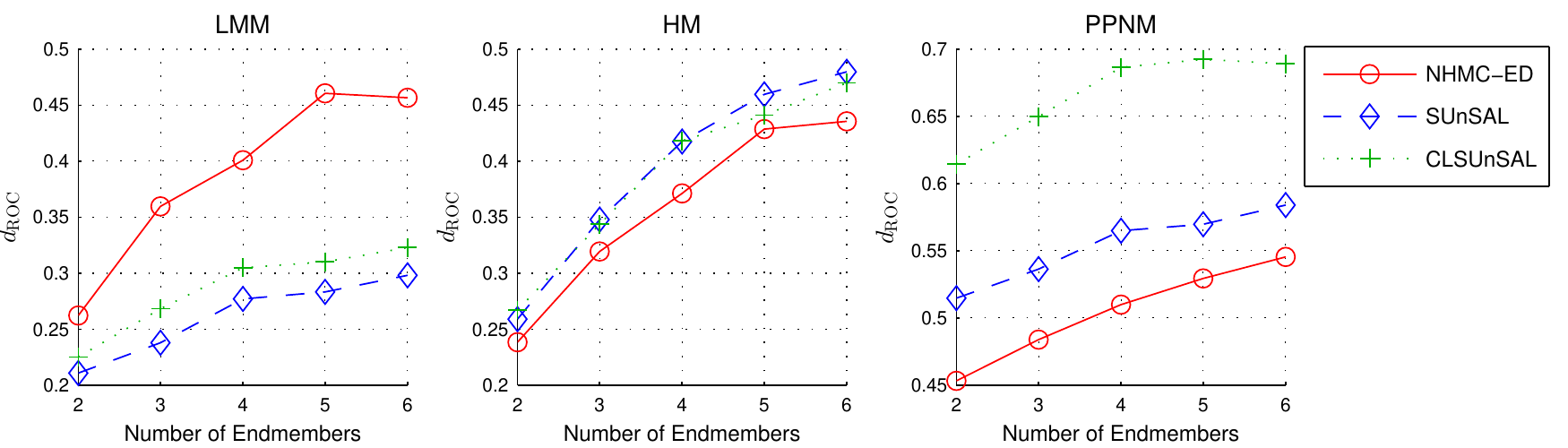}
	\caption{Comparison of the performance of NHMC-ED and SU approaches on synthetic mixtures as a function of different number of endmembers.}
	\label{fig:compPrfnEM}
\end{figure*}
\fi

We also analyze the performance when varying the number of endmembers. Due to  space limitations, we only demonstrate the performance on three different mixture models, LMM, HM, and PPNM. Figure~\ref{fig:compPrfnEM} shows the performance on these three mixing models for the different number of endmembers. While all of the methods tend to perform worse as the number of endmembers increases, the relative performance ranking is consistent throughout.

\subsection{Detection performance with HM model}
\label{sec:ExperimentHapke}
\ifCLASSOPTIONdraftcls
\begin{figure*}[!t]
	\centering
	\subfloat[]{\includegraphics[]{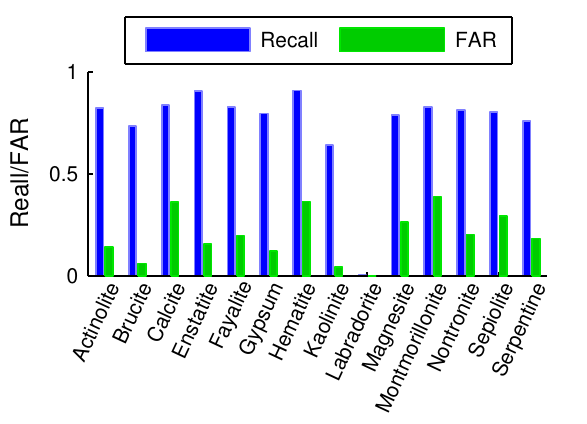}} \hfill
	\subfloat[]{\includegraphics[]{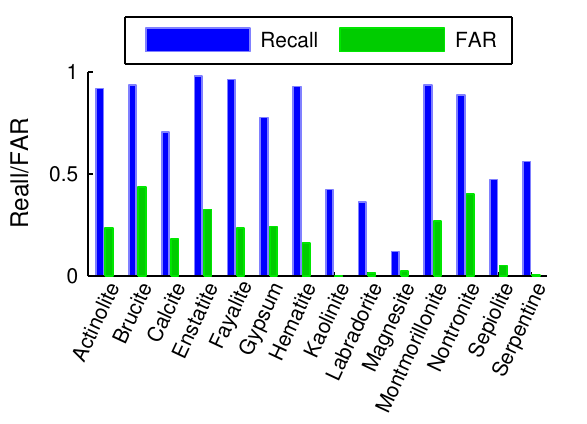}} \hfill
	\subfloat[]{\includegraphics[]{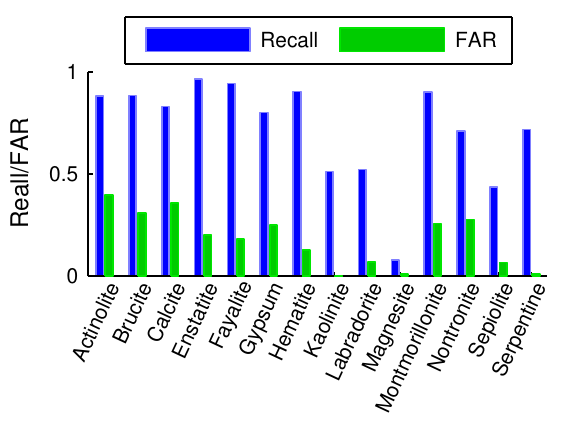}}
	\caption{Recall and FAR per mineral class for (a) NHMC-ED, (b) SUnSAL, and (c) CLSUnSAL.}
	\label{fig:perfclass_syn}
\end{figure*}
\else
\begin{figure*}[!t]
	\centering
	\subfloat[]{\includegraphics[]{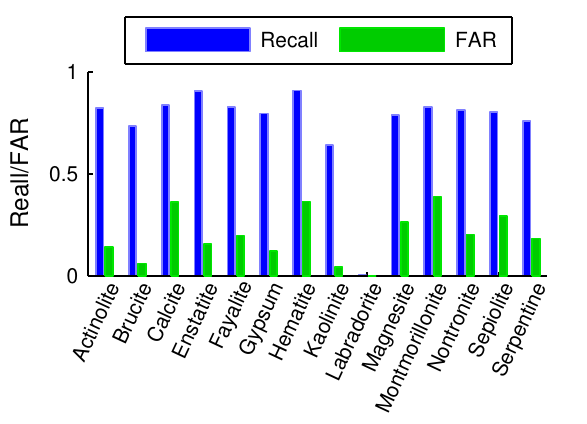}} \hfill
	\subfloat[]{\includegraphics[]{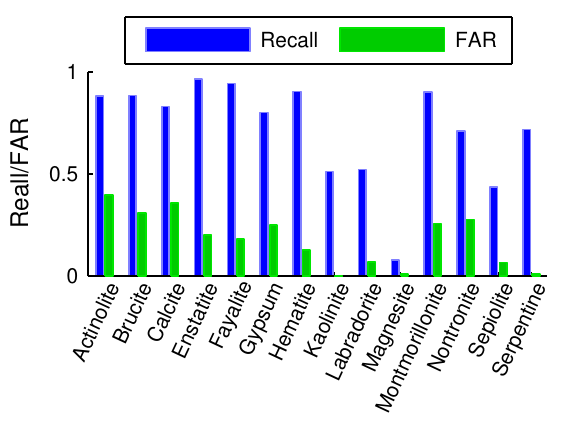}} \hfill
	\subfloat[]{\includegraphics[]{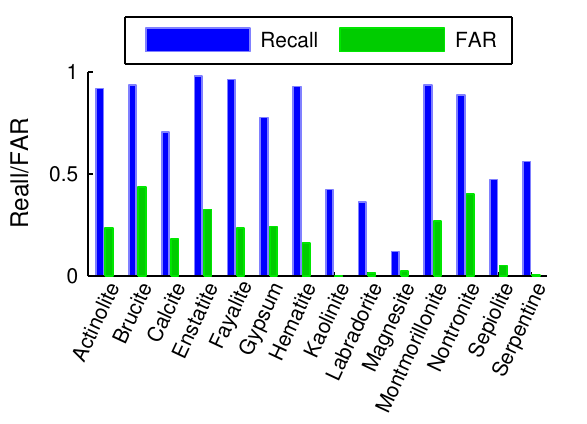}}
	\caption{Recall and FAR per mineral class for (a) NHMC-ED, (b) SUnSAL, and (c) CLSUnSAL.}
	\label{fig:perfclass_syn}
\end{figure*}
\fi
\ifCLASSOPTIONdraftcls
\begin{figure*}[!t]
	\centering
	\includegraphics[]{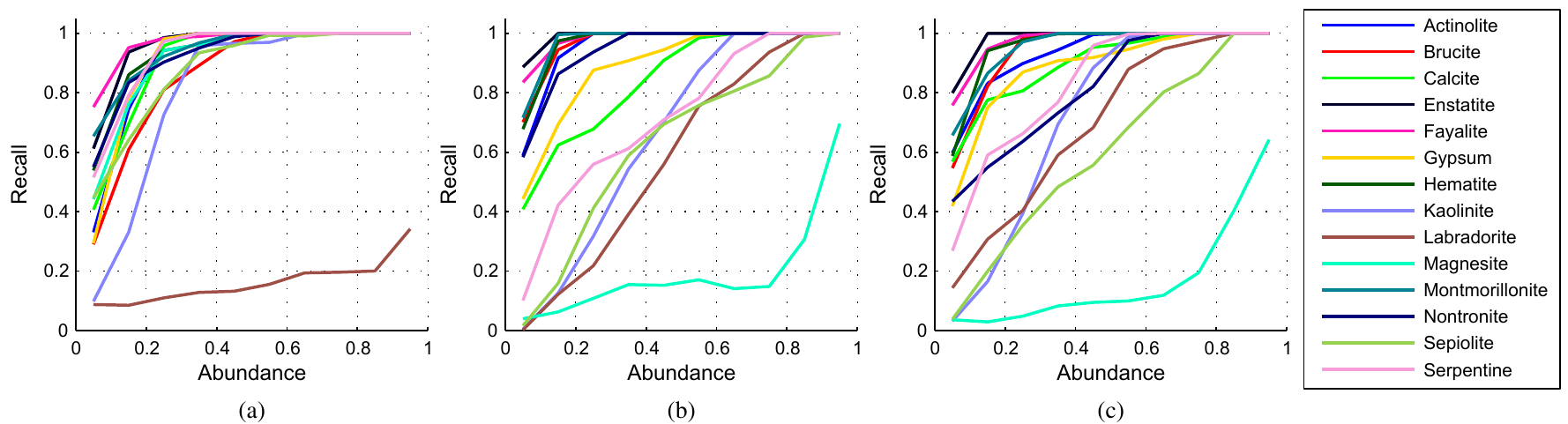}
	\caption{Detection performance w.r.t. abundances of (a) NHMC-ED, (b) SUnSAL, and (c) CLSUnSAL.}
	\label{fig:perfAbuyuki}
\end{figure*}
\else
\begin{figure*}[!t]
	\centering
	\includegraphics[]{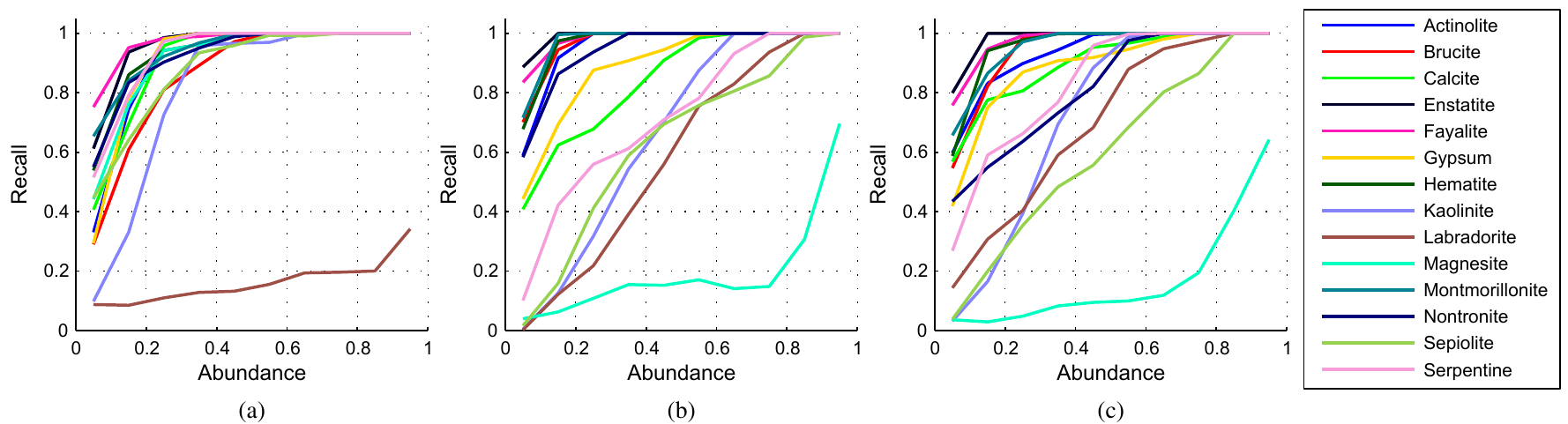}
	\caption{Detection performance w.r.t. abundances of (a) NHMC-ED, (b) SUnSAL, and (c) CLSUnSAL.}
	\label{fig:perfAbuyuki}
\end{figure*}
\fi
We further investigate the per-class detection performance of NHMC-ED and CLSUnSAL on the HM model. Figure~\ref{fig:perfclass_syn} shows the optimal Recall/FAR performance of the three methods. The NHMC-ED performance is significantly more stable across different classes than that of SUnSAL and CLSUnSAL on a class-by-class basis. Figure~\ref{fig:perfAbuyuki} shows the detection performance as a function of the abundance level for each material in the mixture for NHMC-ED, SUnSAL, and CLSUnSAL. According to the figure, the recall quickly becomes larger and surges to one as the abundance increases for the three methods. It is evident that NHMC-ED can recover most endmembers whose abundances are larger than 40\%, except for labradorite. Not only is the average performance of NHMC-ED better than that of SUnSAL and CLSUnSAL, as seen also in Table~\ref{table:prfComp}, but additionally NHMC-ED shows more stable (i.e., similar) performance among the different classes available. This is in contrast to the performance of SUnSAL and CLSUnSAL, which has larger spread (i.e., variability) for different classes, implying that some minerals are easily detected but several other minerals have recalls that surge only slowly. The poor detection performance for labradorite is due to the flatness of its reflectance spectrum, which is difficult to capture by NHMC modeling.

\subsection{Verification of selected features}
\label{sec:ExperimentCharNHMClabels}

We verify that the features identified capture discriminative information from the training library. Figure~\ref{fig:selectedFeatures} shows the selected top 18 features of calcite, kaolinite, montmorillonite, and nontronite aligned with their spectral shapes. In this figure, red marks represent the features selected for the detection of the corresponding mineral class. Although only a small number of features are chosen, we can find significant overlap between the selected semantic features and the discriminative ones determined by geologists~\cite{Tetracorder}. For instance, we successfully detect the absorption band around 2.4 $\mu$m of the calcite's discriminative feature. In addition, we obtained a feature of kaolinite around 2.2 $\mu$m associated with the doublet structure, which is also considered as discriminative by geologists. For montmorillonite and nontronite, we are able to detect the discriminative absorption features around 2.2 $\mu$m and 2.3 $\mu$m, respectively, which are also used to discern these two minerals. Although we show only four mineral classes here due to space limitations, these example observations are representative of other classes and indicate the potential of our method to automatically detect discriminative features.
\ifCLASSOPTIONdraftcls
\begin{figure}[t]
	\centering
	\includegraphics[]{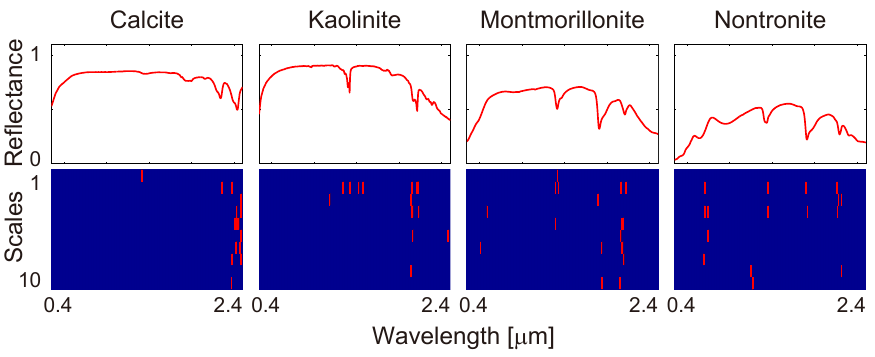}
	\caption{Reflectance spectra and selected NHMC features of calcite, kaolinite, montmorillonite, and nontronite. For each mineral, its reflectance is shown on top and below is its selected NHMC features (in red). The discriminative features of calcite ($2.20-2.40\mu m$), kaolinite ($2.10-2.25\mu m$), montmorillonite ($2.12-2.26\mu m$), and nontronite ($2.25-2.34\mu m$) are clearly selected.   }\label{fig:selectedFeatures}
\end{figure}
\else
\begin{figure}[t]
	\centering
	\includegraphics[]{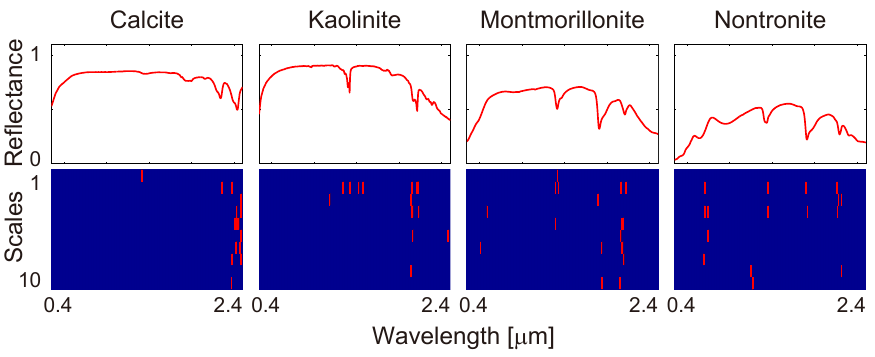}
	\caption{Reflectance spectra and selected NHMC features of calcite, kaolinite, montmorillonite, and nontronite. For each mineral, its reflectance is shown on top and below is its selected NHMC features (in red). The discriminative features of calcite ($2.20-2.40\mu m$), kaolinite ($2.10-2.25\mu m$), montmorillonite ($2.12-2.26\mu m$), and nontronite ($2.25-2.34\mu m$) are clearly selected.   }\label{fig:selectedFeatures}
\end{figure}
\fi

\section{Experimental results on real data}
\label{sec:experiment2}
We apply our NHMC-ED to a real hyperspectral data set. The hyperspectral image (HSI) used in this experiment was acquired by the airborne visible and infrared Spectrometer (AVIRIS)~\cite{Vane1993RSE} on the Cuprite mining site in Nevada in 1995. We used a subset of the HSI with the size of $614\times 750$ that is distributed with Tetracorder.\footnote{A sample AVIRIS data is available for download~\cite{AVIRISDL}.} Figure~\ref{fig:cupriteImage}(a) shows a pseudo-RGB image for the HSI using three bands (24, 16, and 12). 

In this experiment, the mineral maps generated by the Tetracorder (v.\ 4.4)~\cite{Swayze2014cuprite} are considered to be the ground truth. The Tetracorder is an expert system to map the distribution of the minerals that exist in the spectral library by matching each observed spectral signature with individual signatures in the library based only on their hand-picked discriminative features. The Tetracorder outputs a collection of images, each of which shows the matching scores for a given class corresponding to a mineral or a mineral mixture. An ``unknown'' label is assigned to pixels that do not have sufficiently high scores for any of the signatures in the Tetracorder reference library.
\ifCLASSOPTIONdraftcls
	\begin{figure*}[t]
		\centering
		\includegraphics[]{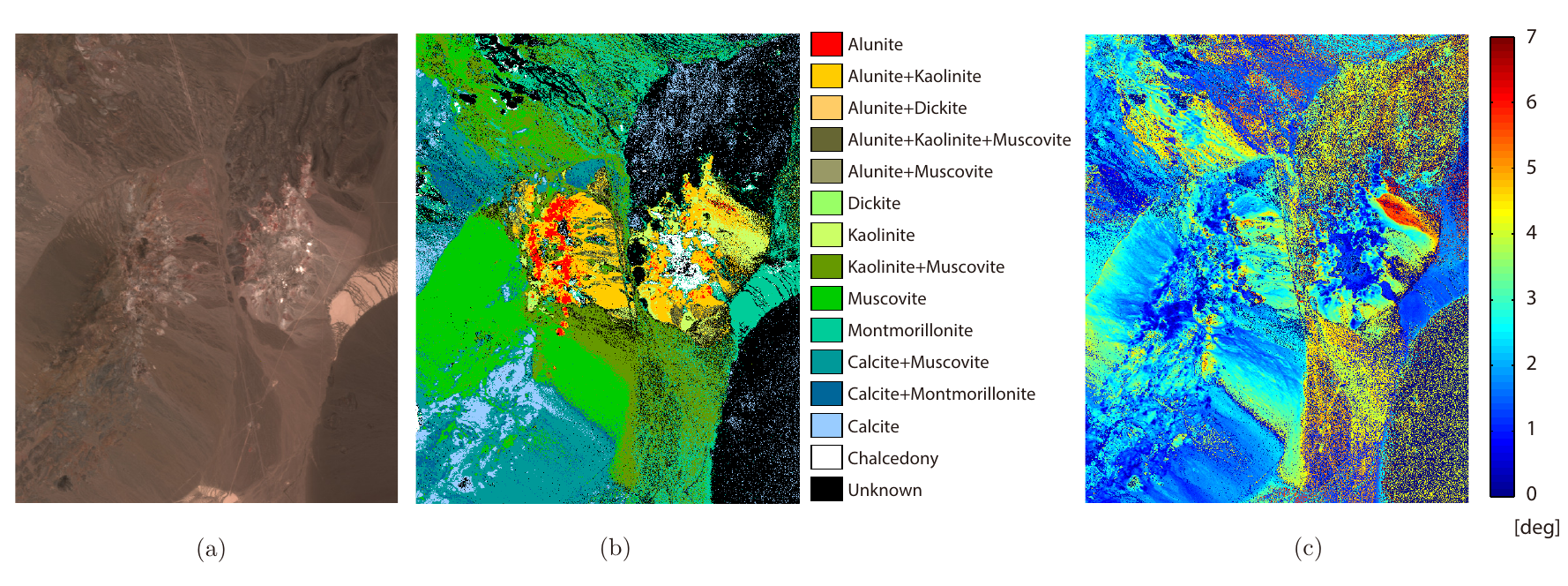}
		\caption{Cuprite hyperspectral image used in our experiments. (a) pseudo-RGB image of the HSI used for testing,  (b) mapping results of the Tetracorder, and (c) $\mathrm{NS}$ map.}
		\label{fig:cupriteImage}
	\end{figure*}
\else
	\begin{figure*}[t]
		\centering
		\includegraphics[]{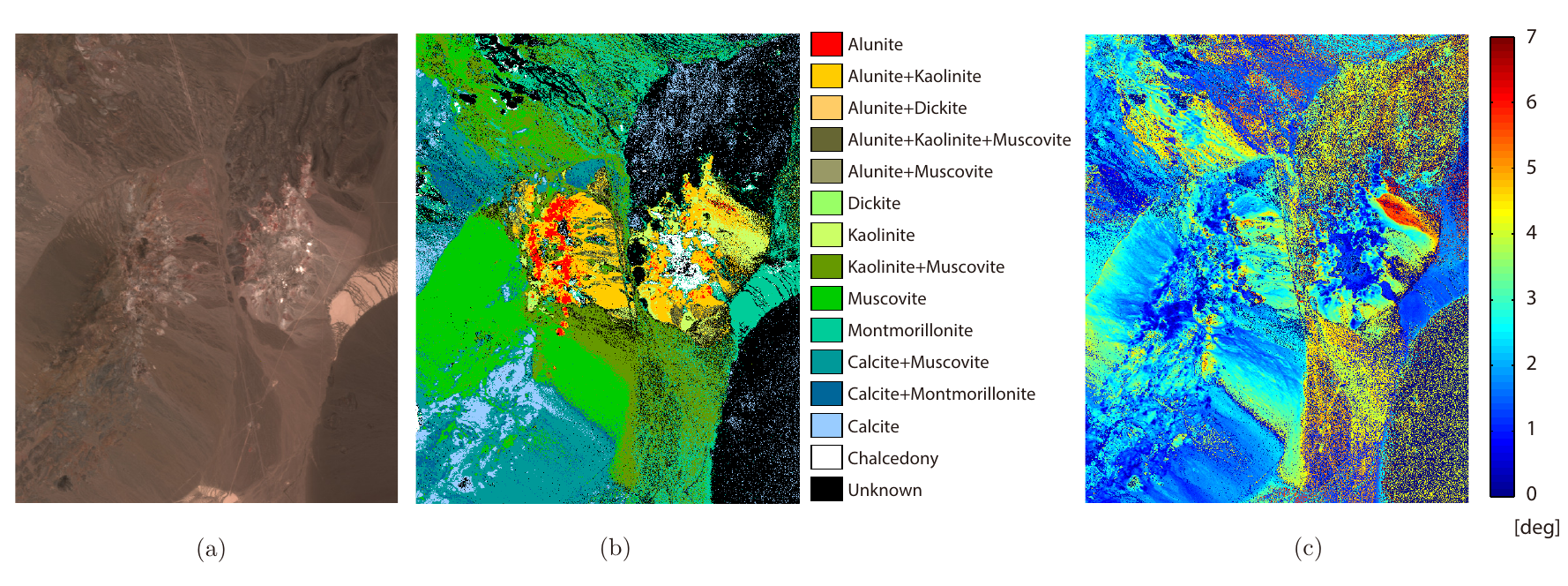}
		\caption{Cuprite hyperspectral image used in our experiments. (a) Pseudo-RGB image of the HSI used for testing, (b) Mapping results of the Tetracorder, and (c) $\mathrm{NS}$ map.}
		\label{fig:cupriteImage}
	\end{figure*}
\fi
\ifCLASSOPTIONdraftcls
\begin{figure*}[]
	\centering
	\includegraphics[]{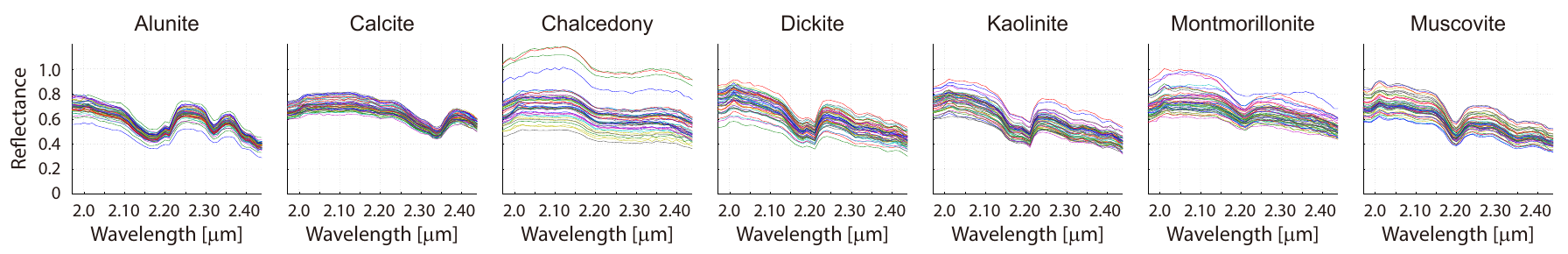}
	\caption{Endmembers extracted from the image.}
	\label{fig:cupriteEMs}
\end{figure*}
\else
\begin{figure*}[]
	\centering
	\includegraphics[]{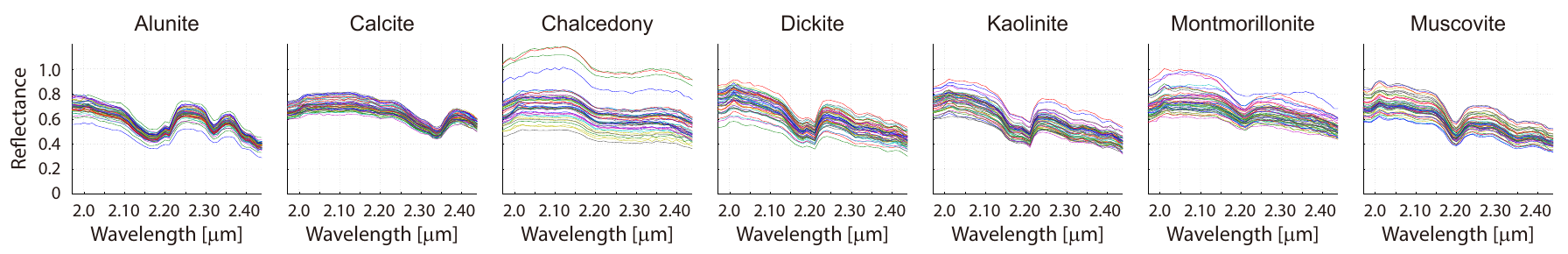}
	\caption{Endmembers extracted from the image.}
	\label{fig:cupriteEMs}
\end{figure*}
\fi

We focus on 48 bands (band 170-217) in the short wave infrared (SWIR) wavelength region. We visually examined the scores of the Tetracorder and the spectral shape at each pixel. We perform hard thresholding of the score at 10 to obtain ground truth labels for each pixel because spectral signatures featuring scores lower than 10 do not resemble the reference spectra. Figure~\ref{fig:cupriteImage}(b) shows the distribution of these minerals in the HSI, showing that the scene is mainly composed of seven minerals: alunite, calcite, chalcedony, dickite, kaolinite, montmorillonite, and muscovite. Note that some of the classes selected by the Tetracorder are merged into one class. Figure~\ref{fig:cupriteImage}(c) shows the map of $\mathrm{NS}$. The non-linearity scores are relatively high at many pixels and therefore, it is anticipated that NHMC-ED performs better than SU approaches.

The spectral library for the experiment is created by extracting pixels in the image that are considered to be sufficiently pure according to the Tetracorder scores; Figure~\ref{fig:cupriteEMs} shows the spectral signatures for the chosen endmembers. The reason why we use image endmembers instead of spectral samples from the U.S. Geological Survey (USGS) spectral library~\cite{splib06} is that each of the mineral classes has an insufficient number of samples in the USGS library. Since the NHMC model uses the statistics of the spectral signatures in each class, a moderate number of samples is necessary to detect discriminative features.

After constructing the library and setting up the training feature labels for the pixels in the image, we apply NHMC-ED and SUnSAL. The parameters of NHMC-ED are learned on the spectral library. The attenuations of library spectra range on a grid from 0.1 and 1 with a step size of 0.1, as done in the last section. To measure the detection performance of SUnSAL and CLSUnSAL, we also apply thresholding to SUnSAL's (and CLSUnSAL's) vector of estimated abundances. We optimize the parameters of NHMC ($k$ and the number of features) and SUnSAL and CLSUnSAL (the trade-off parameter and the thresholding value) so that the performance is maximized in terms of $d_\text{ROC}$ as done in the experiment on the simulated data. In this experiment, we searched the optimal number of NHMC states over $[2,3,\ldots,8]$. The other parameter range is same as the experiments on simulated data. The best performance of NHMC-ED was obtained at $k=5$ and 21 for the number of features. The number of scales used for the wavelet transform the proposed method is set to ten. For SUnSAL, the trade-off parameter is set to 0.0 and the additional hard thresholding with the threshold value $0.2$ is applied. The zero value of the trade-off parameter indicates that the performance of endmember identification for SUnSAL with thresholding is achieved without $\ell_1$-regularization, which is coherent with the result in~\cite{Itoh2016IGARSS}. The performance of the CLSUnSAL is maximized when the trade-off parameter and the threshold value are equal to 0.01 and 0.2.%
\ifCLASSOPTIONdraftcls
\begin{figure*}[]
	\centering
	\includegraphics[width=469pt]{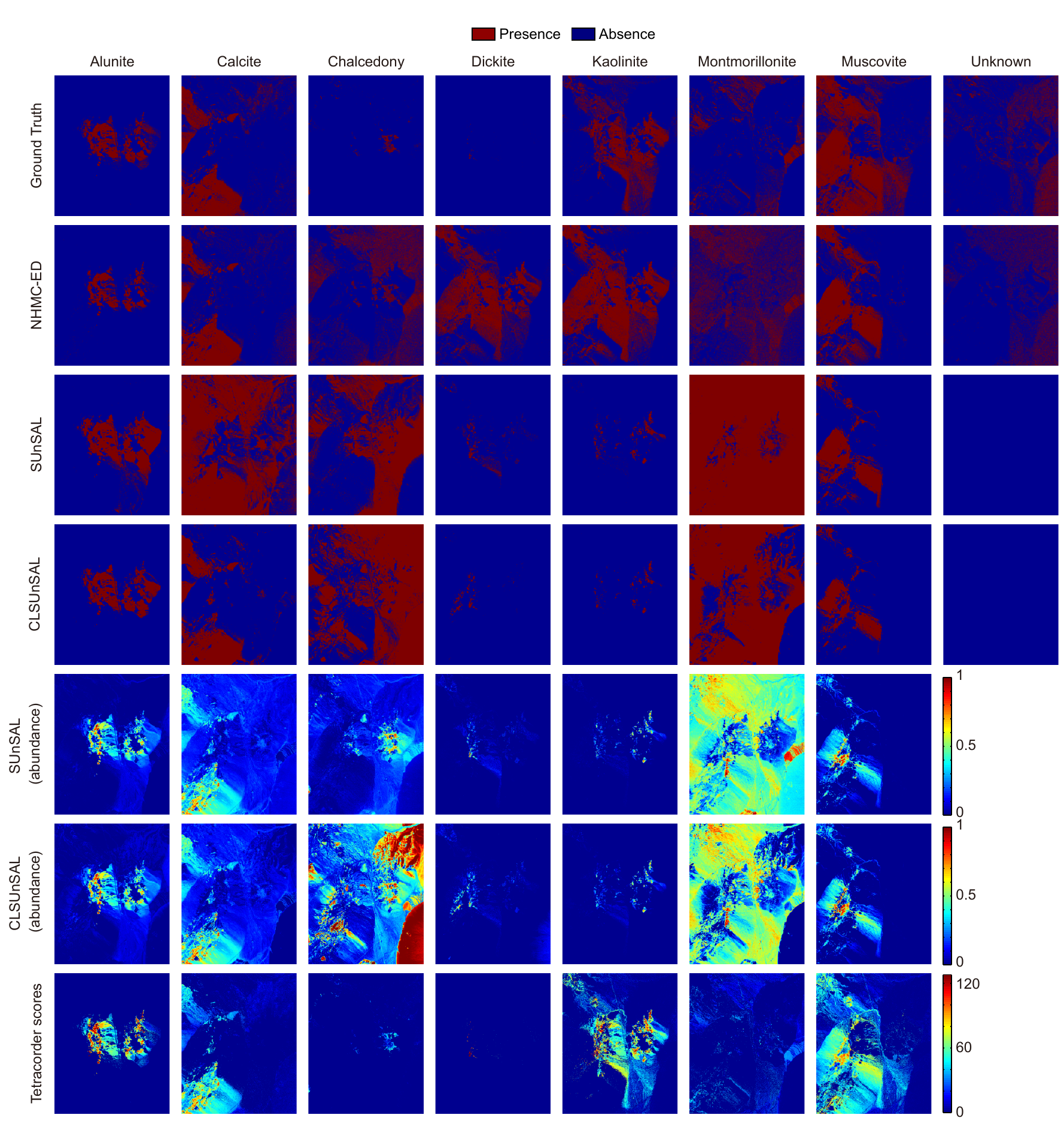}
	\caption{Comparison of the mapping results and abundance estimates (except for the ``unknown'' class) for the minerals of interest obtained by the Tetracorder, NHMC-ED, SUnSAL, and CLSUnSAL.}
	\label{fig:cupriteMapping}
\end{figure*}
\else
\begin{figure*}[]
	\centering
	\includegraphics[]{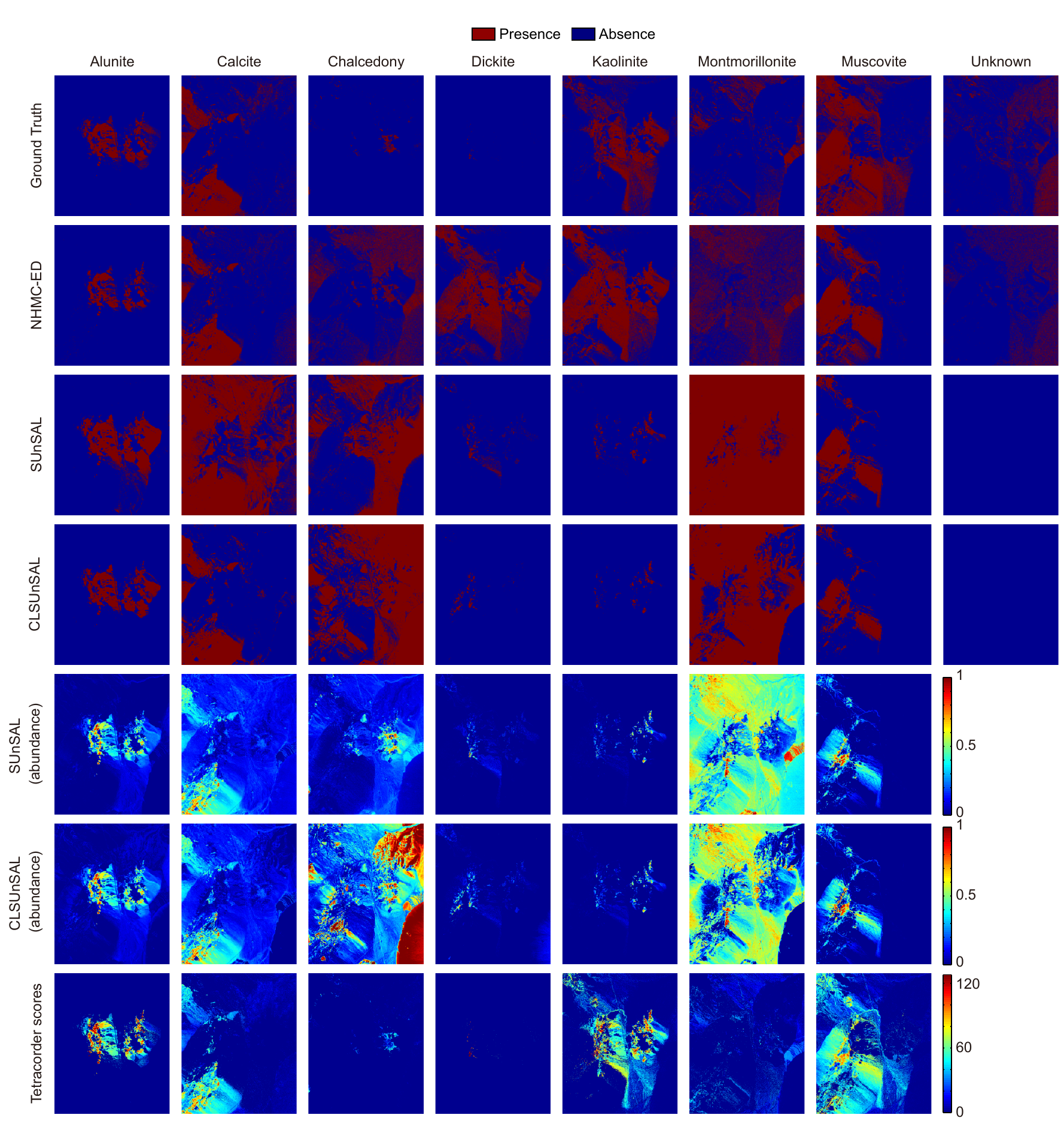}
	\caption{Comparison of the mapping results and abundance estimates (except for the ``unknown'' class) for the minerals of interest obtained by the Tetracorder, NHMC-ED, SUnSAL, and CLSUnSAL.}
	\label{fig:cupriteMapping}
\end{figure*}
\fi

We measure the detection performance by using the ground truth labels obtained by the Tetracorder as a reference. 
The first four rows of Figure~\ref{fig:cupriteMapping} show a comparison between the mineral mappings obtained by the Tetracorder, NHMC-ED, SUnSAL, and CLSUnSAL. 
The red (blue) pixels in the Tetracorder maps represent scores higher (lower) than the threshold value set above. For NHMC-ED, SUnSAL, and CLSUnSAL, red pixels represent pixels with detection of the particular endmember. 

From visual inspection, NHMC-ED improves the detection performance of six mineral classes, with dickite being the sole exception. Our method allows for the detection of abundant kaolinite in this scene; furthermore, the distributions of montmorillonite obtained by NHMC-ED more closely resemble the ground truth than those obtained by SUnSAL and CLSUnSAL, although the false alarms for dickite is high for NHMC. We conjecture that these false alarms appear because dickite has high spectral similarity to kaolinite. In contrast, SUnSAL and CLSUnSAL tend to often falsely detect montmorillonite, and chalcedony. Those minerals have relatively flat spectra and appear to have been detected to compensate for a smooth distortion that is present in the spectra over the wavelength region being considered.  

Figure~\ref{fig:perfclass} shows the Recall/FAR performance of the three methods. The NHMC-ED performance is significantly more stable across different classes than that of SUnSAL and CLSUnSAL on a class-by-class basis. Nonetheless, the average performances are comparable: the average recall and FAR of NHMC-ED are 70\% and 19\% respectively, those of SUnSAL are 66\% and 16\% respectively, and those of CLSUnSAL are 67\% and 20\%, respectively We argue that NHMC-ED is preferable over the others because of the stability of the performance for different classes. The stability was also claimed in the experiments on simulated data. Note that unknown pixels are also counted for the computation of FAR; even when those pixels are excluded, the average FARs for those methods only change less than 1\%.

\ifCLASSOPTIONdraftcls
	\begin{figure}[t]
		\centering
		\includegraphics[]{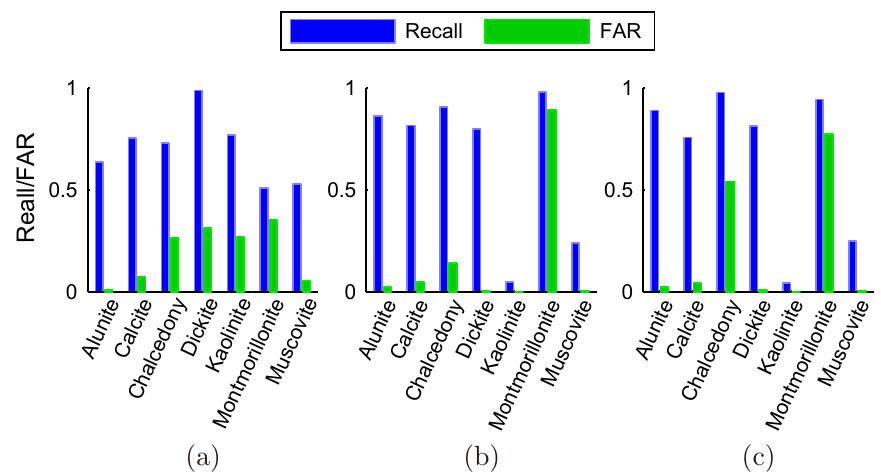}
		\caption{Recall and FAR per mineral class for (a) NHMC-ED, (b) SUnSAL, and (c) CLSUnSAL.}
		\label{fig:perfclass}
	\end{figure}
\else
	\begin{figure}[t]
		\centering
		\includegraphics[]{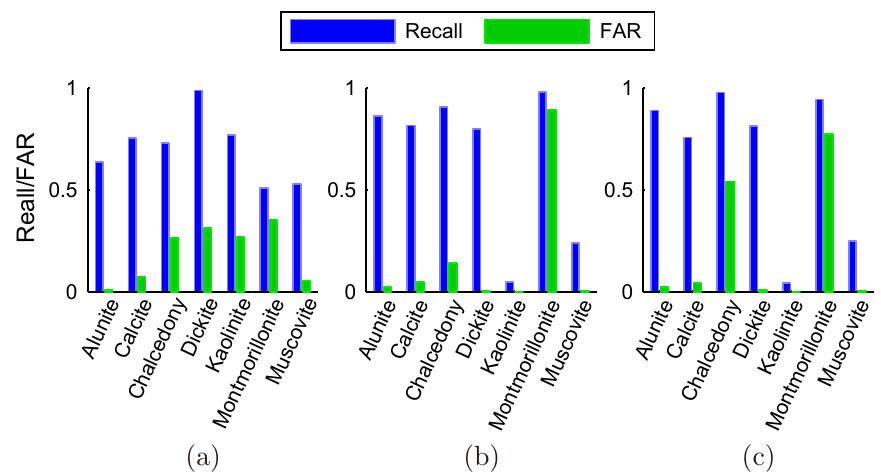}	
		\caption{Recall and FAR per mineral class for (a) NHMC-ED, (b) SUnSAL, and (c) CLSUnSAL.}
		\label{fig:perfclass}
	\end{figure}
\fi

Next, we investigate the challenges of SUnSAL and CLSUnSAL. The fifth, sixth, and seventh rows in Fig.~\ref{fig:cupriteMapping} show the estimated abundances by SUnSAL, those by CLSUnSAL and the original Tetracorder scores, respectively. Comparing these three rows, we can find moderate consensus between the SUnSAL's or CLSUnSAL's estimated abundances and the Tetracorder scores, except for kaolinite and montmorillonite. The SUnSAL's estimated abundance maps of montmorillonite look very different; nonetheless, after applying a higher threshold level in SUnSAL, its detection result is much closer to the ground truth in the first row. This finding indicates that proper thresholding on the estimated abundances could improve estimation accuracy. Since the same threshold value is applied to the different mineral classes in this experiment, the detection performance could potentially be improved by setting different threshold values for individual mineral classes. However, this would exponentially increase the complexity of the parameter space. In contrast, it seems difficult to improve the detection performance of CLSUnSAL just by changing the threshold values.

We also explore the mapping accuracy for pixels classified as unknown in the ground truth. This ``unknown'' label is given to the pixels with no label or labeled with mineral classes outside the aforementioned seven minerals detected by the Tetracorder. In NHMC-ED, the ``unknown'' label is assigned to the pixels for which all detectors return negative labels; for SUnSAL and CLSUnSAL, the ``unknown'' label is assigned to the pixels for which all the estimated abundances are below the threshold value. Recall that the threshold value is optimized to maximize the performance in all seven classes, excluding the unknown class. The last column in Fig.~\ref{fig:cupriteMapping} shows the membership of the ``unknown'' class for the three aforementioned methods. NHMC-ED produces a map that is visually similar to that of the ground truth, while SUnSAL and CLSUnSAL do not identify any ``unknown'' pixels. One of the most notable unknown areas is around the lower right region in the image. For this region, SUnSAL tends to assign montmorillonite and CLSUnSAL seems to assign chalcedony instead. This demonstrates another disadvantage of these two methods; they need all endmembers present in the scene to be part of the dictionary in order to obtain successful performance. As observed, missing endmembers are often compensated in sparsity-based methods by selecting other (incorrect) endmembers. On the other hand, NHMC-ED was able to cope with the ``unknown'' class as well as Tetracorder does.

Finally, we investigate the performance of the three methods as a function of the degree of nonlinearity in the mixture, using the same procedure as in Section~\ref{sec:experiment1}. We use the same nonlinearity score~\eqref{eq:NS}, with $\mathbf{W}$ being composed of all the endmembers for the mineral classes present in the pixel. The pixels in the testing set are clustered into groups possessing different levels of nonlinearity (NS = 0-1\degree, 1-2\degree, 2-3\degree, etc.). The performance of the three algorithms is then evaluated for each nonlinearity level. Figure~\ref{fig:cupriteResultsNonlinearity} shows the average value of $d_{\mathrm{ROC}}$ as a function of $\mathrm{NS}$. As seen in Table~\ref{table:prfComp} in the experiments on simulated data, NHMC-ED outperforms SUnSAL and CLSUnSAL in the presence of a sufficiently strong mixture nonlinearity. 
\ifCLASSOPTIONdraftcls
	\begin{figure}[t]
		\centering
		\includegraphics[]{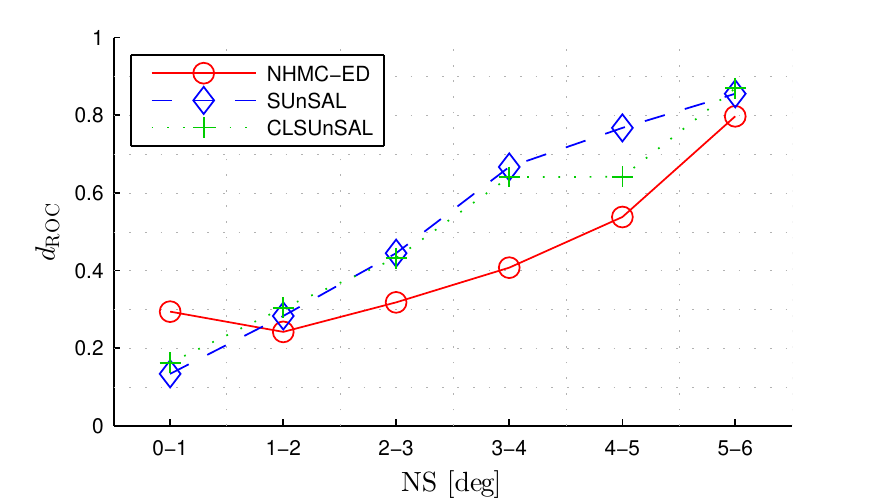}
		\caption{Comparison of the performance of NHMC-ED, SUnSAL, and CLSUnSAL on the AVIRIS HSI as a function of the nonlinearity score.}
		\label{fig:cupriteResultsNonlinearity}
	\end{figure}
\else
	\begin{figure}[t]
		\centering
		\includegraphics[]{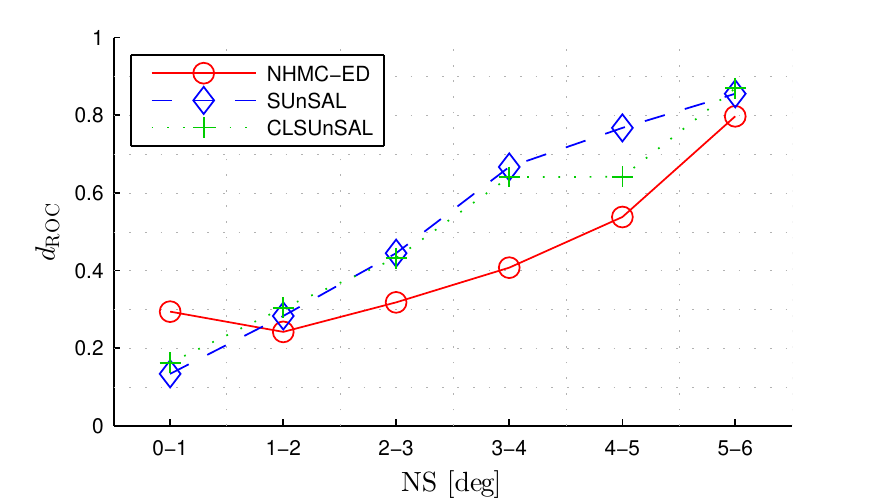}
		\caption{Comparison of the performance of NHMC-ED, SUnSAL, and CLSUnSAL on the AVIRIS HSI as a function of the nonlinearity score.}
		\label{fig:cupriteResultsNonlinearity}
	\end{figure}
\fi
\section{Conclusion and Future works}
\label{sec:conclusion}
In this paper, we have presented NHMC-ED, a new method to detect endmembers present in nonlinear mixtures using a semantic representation for hyperspectral signals. NHMC-ED uses the semantic representation that is obtained from NHMC models and a series of detectors that are designed to determine the presence of individual materials. In each detector, a modified CMI feature selection method is adopted for unmixing tasks. One of the advantages of NHMC-ED is that it is agnostic to the mixing model present in the observations. Experimental results show that NHMC-ED exploits discriminative features similar to those determined by experts, and that NHMC-ED can be a promising detection method for highly nonlinear mixing scenarios. Our results also reiterate the potential of the NHMC-based semantic representations for encoding scientific information. 

One possible future direction would be the combination of NHMC-ED with prior de-noising on wavelet coefficients using soft-thresholding~\cite{Donoho1995} or hidden Markov tree modeling~\cite{Crouse1998} to achieve robustness to noise. Since the wavelet coefficients are strongly affected by the presence of noise, it is expected that the performance of our proposed method will be deteriorated as noise increases, implying the necessity of prior de-noising. Another future work would be the investigation of the properties of the library critical for the performance of NHMC-ED. Since our semantic representation is a non-linear function of the spectrum, the properties of the library to assess the performance of SU (such as coherence) may not be useful, pointing to the need for different metrics to determine well-conditioned libraries.

\section*{Acknowledgment}
The authors would like to thank Jos\'{e} M. Bioucas-Dias and M\'{a}rio A. T. Figueiredo for making their Matlab code for SUnSAL available online, Takahiro Hiroi and the team at Brown University for making the RELAB spectral database online, and the authors of~\cite{Swayze2014cuprite} for making the Tetracorder software and the Cuprite image available online.


\ifCLASSOPTIONcaptionsoff
  \newpage
\fi
\ifCLASSOPTIONdraftcls
\newpage
\fi



\bibliographystyle{IEEEtran_yuki}
\bibliography{library_corr,custom}
\end{document}